\documentclass{article}

\PassOptionsToPackage{numbers, sort&compress}{natbib}
\usepackage[final]{nips_2016}

\usepackage[utf8]{inputenc} 
\usepackage[T1]{fontenc}    
\usepackage{hyperref}       
\usepackage{url}            
\usepackage{booktabs} 
\usepackage{amsfonts} 
\usepackage{nicefrac}   
\usepackage{microtype}
\usepackage{amsmath}
\usepackage{graphicx}
\usepackage{xcolor}

\usepackage{amsmath}
\usepackage{amssymb}
\usepackage{amsthm}

\newtheorem{lemma}{Lemma}
\newtheorem{theorem}{Theorem}
\newtheorem{proposition}{Proposition}
\newtheorem{defn}{Definition}
\newenvironment{definition}{\begin{defn}\em}{\end{defn}}
\newtheorem{exampl}{Example}
\newenvironment{example}{\begin{exampl}\em}{\end{exampl}}

\newcommand{\FOtwo}{\mathit{FO}^2}
\newcommand{\StwoFOtwo}{\mathit{S}^2\mathit{FO}^2}
\newcommand{\StwoRU}{\mathit{S}^2\mathit{RU}}

\newcommand{\pop}[1]{\Delta_{#1}}
\newcommand{\lv}[1]{#1}
\newcommand{\ind}[1]{#1}
\newcommand{\const}[1]{#1}
\newcommand{\pred}[1]{\ensuremath{\mathsf{#1}}}
\newcommand{\true}{\ensuremath{\mathsf{True}}}
\newcommand{\false}{\ensuremath{\mathsf{False}}}
\newcommand{\rules}{\ensuremath{\mathcal{R}}}
\newcommand{\rulesplus}{\ensuremath{\mathcal{R^D}}}
\newcommand{\predsyms}[1]{\ensuremath{\mathcal{F}{(#1)}}}
\newcommand{\wfomc}{\ensuremath{\mathrm{WFOMC}}}

\newlength{\myparskip}
\newlength{\myparindent}
\newlength{\mytopsep}
\title{New Liftable Classes for \\ First-Order Probabilistic Inference}

\author{
  Seyed Mehran Kazemi \\
  The University of British Columbia\\
  \texttt{smkazemi@cs.ubc.ca}  \\
  \And
  Angelika Kimmig \\
  KU Leuven \\
\texttt{angelika.kimmig@cs.kuleuven.be} \\
\And
Guy Van den Broeck \\
University of California, Los Angeles \\
\texttt{guyvdb@cs.ucla.edu} \\
\And
David Poole \\
The University of British Columbia \\
\texttt{poole@cs.ubc.ca}
}

\begin{document}

\maketitle

\begin{abstract}
Statistical relational models provide compact encodings of probabilistic dependencies in relational domains, but result in highly intractable graphical models. The goal of lifted inference is to carry out probabilistic inference without needing to reason about each individual separately, by instead treating exchangeable, undistinguished objects as a whole. In this paper, we study the \emph{domain recursion} inference rule, which, despite its central role in early theoretical results on domain-lifted inference, has later been believed redundant. We show that this rule is more powerful than expected, and in fact significantly extends the range of models for which lifted inference runs in time polynomial in the number of individuals in the domain. This includes an open problem called S4, the symmetric transitivity model, and a first-order logic encoding of the birthday paradox. We further identify new classes $\StwoFOtwo$ and $\StwoRU$ of domain-liftable theories, which respectively subsume $\FOtwo$ and \emph{recursively unary theories}, the largest classes of domain-liftable theories known so far, and show that using domain recursion can achieve exponential speedup even in theories that cannot fully be lifted with the existing set of inference rules. 
\end{abstract}

\section{Introduction}
Statistical relational learning (SRL) \citep{StarAI-Book} aims at unifying logic and probability for reasoning and learning in noisy domains, described in terms of individuals (or objects), and the relationships between them. Statistical relational models \citep{getoor2007introduction}, or template-based models~\citep{Koller:2009} extend Bayesian and Markov networks with individuals and relations, and compactly describe probabilistic dependencies among them. These models encode exchangeability among the objects: individuals that we have the same information about are treated similarly.

A key challenge with SRL models is the fact that they represent highly intractable, densely connected graphical models, typically with millions of random variables. The aim of \emph{lifted inference} \citep{Poole:2003} is to carry out probabilistic inference without needing to reason about each individual separately, by instead treating exchangeable, undistinguished objects as a whole. 
Over the past decade, a large number of lifted inference rules have been proposed~\citep{De:2005,Milch:2008,Poole:2011,Choi:2011,jha2010lifted,PTP,Van:2011,van2014skolemization}, often providing exponential speedups for specific SRL models. These basic exact inference techniques have applications in (tractable) lifted learning \citep{van2015lifted}, where the main task is to efficiently compute \emph{partition functions}, and in variational and over-symmetric approximations~\citep{VdBUAI12,venugopal2014evidence}. Moreover, they provided the foundation for a rich literature on approximate lifted inference and learning~\cite{singla2008lifted,kersting2009counting,niepert2012markov,bui2013automorphism,VenugopalNIPS14,kopp2015lifted,ahmadi2012lifted,jernite2015fast}.

The theoretical study of lifted inference began with the complexity notion of \emph{domain-lifted} inference~\citep{gvdb2011completeness} (a concept similar to data complexity in databases). Inference is domain-lifted when it runs in time polynomial in the number of individuals in the domain. By identifying \emph{liftable classes} of models, guaranteeing domain-lifted inference, one can characterize the theoretical power of the various inference rules. 
For example, the class $\FOtwo$, encoding dependencies among pairs of individuals (i.e., two logical variables), is liftable~\cite{van2014skolemization}. \citet{LRC2CPP} introduce a liftable class called \emph{recursively unary}, capturing hierarchical simplification rules. \citet{beame2015symmetric} identify liftable classes of probabilistic database queries.
Such results elevate the specific inference rules and examples to a general principle, and bring lifted inference in line with complexity and database theory~\cite{beame2015symmetric}.

This paper studies the \emph{domain recursion} inference rule, which applies the principle of induction on the domain size. The rule makes one individual $\const{A}$ in the domain explicit. Afterwards, the other inference rules simplify the SRL model up to the point where it becomes identical to the original model, except the domain size has decreased.
Domain recursion was introduced by \citet{gvdb2011completeness} and was central to the proof that $\FOtwo$ is liftable.
However, later work showed that simpler rules suffice to capture $\FOtwo$~\cite{taghipour2013completeness}, and the domain recursion rule was forgotten.

We show that domain recursion is more powerful than expected, and can lift models that are otherwise not amenable to domain-lifted inference. 
This includes an open problem by \citet{beame2015symmetric}, asking for an inference rule for a logical sentence called S4. It also includes the symmetric transitivity model, and an encoding of the birthday paradox in first-order logic. There previously did not exist any efficient algorithm to compute the partition function of these SRL models, and we obtain exponential speedups.
Next, we prove that domain recursion supports its own large \emph{classes of liftable models} $\StwoFOtwo$ subsuming $\FOtwo$, and $\StwoRU$ subsuming recursive unary.
All existing exact lifted inference algorithms (e.g., \citep{PTP,Van:2011,LRC2CPP}) resort to grounding the theories in $\StwoFOtwo$ or $\StwoRU$ that are not in $\FOtwo$ or recursively unary, and require time exponential in the domain size. 

These results will be established using the weighted first-order model counting (WFOMC) formulation of SRL models~\citep{Van:2011}. WFOMC is close to classical first-order logic, and it can encode many other SRL models, including Markov logic \citep{Richardson:2006aa}, parfactor graphs~\cite{Poole:2003}, some probabilistic programs \citep{DeRaedt:2007}, relational Bayesian networks \citep{Jaeger:1997}, and probabilistic databases \citep{suciu2011probabilistic}. It is a basic specification language that simplifies the development of lifted inference algorithms \citep{PTP,Van:2011,beame2015symmetric}.

\section{Background and Notation}
A \textbf{population} is a set of constants denoting individuals (or objects). A \textbf{logical variable (LV)} is typed with a population. We represent LVs with lower-case letters, constants with upper-case letters, the population associated with a LV $\lv{x}$ with $\pop{x}$, and its cardinality with $|\pop{x}|$.That is, a population $\pop{x}$ is a set of constants $\{\const{X}_1,\ldots,\const{X}_n\}$, and we use $\lv{x}\in\pop{x}$ as a shorthand for instantiating $\lv{x}$ with one of the $\const{X}_i$.
A \textbf{parametrized random variable (PRV)} is of the form $\pred{F}(t_1, \dots, t_k)$ where $\pred{F}$ is a predicate symbol and each $t_i$ is a LV or a constant. A \textbf{unary} PRV contains exactly one LV and a \textbf{binary} PRV contains exactly two LVs. 
A \textbf{grounding} of a PRV is obtained by replacing each of its LVs $\lv{x}$ by one of the individuals in $\pop{x}$.

A \textbf{literal} is a PRV or its negation.
A \textbf{formula} $\varphi$ is a literal, a disjunction $\varphi_1\vee\varphi_2$ of formulas, a conjunction $\varphi_1\wedge\varphi_2$ of formulas, or a quantified formula $\forall\lv{x}\in\pop{x}:\varphi(\lv{x})$ or $\exists\lv{x}\in\pop{x}:\varphi(\lv{x})$ where $\lv{x}$ appears in $\varphi(\lv{x})$.
A \textbf{sentence} is a formula with all LVs quantified. 
A \textbf{clause} is a formula not using conjunction. 
A \textbf{theory} is a set of sentences.
A theory is \textbf{clausal} if all its sentences are clauses. 
An \textbf{interpretation} is an assignment of values to all ground PRVs in a theory. 
An interpretation~$I$ is a \textbf{model} of a theory~$T$, $I\models T$, if given its value assignments, all sentences in $T$ evaluate to \true.

Let $\predsyms{T}$ be the set of predicate symbols in theory $T$, and $\Phi: \predsyms{T}\rightarrow \mathbb{R}$ and $\overline{\Phi}: \predsyms{T}\rightarrow \mathbb{R}$ be two functions 
that map each predicate~$\pred{F}$ to weights for ground PRVs~$\pred{F}(\const{C}_1,\ldots,\const{C}_k)$ assigned $\true$ and $\false$ respectively.
For an interpretation $I$ of $T$, let $\psi^{True}$ be the set of ground PRVs assigned \true, and $\psi^{False}$ the ones assigned \false. The weight of $I$ is given by $\omega(I)=\prod_{ \pred{F}(\const{C}_1,\ldots,\const{C}_k)\in \psi^{True}} \Phi(\pred{F}) \cdot \prod_{\pred{F}(\const{C}_1,\ldots,\const{C}_k) \in \psi^{False}} \overline{\Phi}(\pred{F})$. 
Given a theory $T$ and two functions $\Phi$ and $\overline{\Phi}$, the \textbf{weighted first-order model count (WFOMC)} of the theory given $\Phi$ and $\overline{\Phi}$ is:
$\wfomc(T | \Phi, \overline{\Phi}) = \sum_{I \models T} \omega(I)$.

In this paper, we assume that all theories are clausal and do not contain existential quantifiers. The latter can be achieved 
using the skolemization procedure of \citet{van2014skolemization}, which transforms a theory $T$ with existential quantifiers into a theory $T'$ without existential quantifiers that has the same weighted model count, in time polynomial in the size of $T$. 
That is, our theories are sets of finite-domain, function-free first-order clauses whose LVs are all universally quantified (and typed with a population). 
Furthermore, when a clause mentions two LVs $\lv{x_1}$ and $\lv{x_2}$ with the same  population $\pop{x}$, or a LV $\lv{x}$ with population $\pop{x}$ and a constant~$\const{C}\in\pop{x}$, we assume they refer to different individuals.\footnote{Equivalently, we can disjoin $\lv{x_1}\neq\lv{x_2}$ or $\lv{x}\neq\const{C}$ to the clause.}

\begin{example}
Consider the theory $ \forall \lv{x} \in \pop{x}: \neg \pred{Smokes}(\lv{x}) \vee \pred{Cancer}(\lv{x})$ having only one clause and assume $\pop{x} = \{\const{A}, \const{B}\}$. The assignment $\pred{Smokes}(\const{A})=\true, \pred{Smokes}(\const{B})=\false, \pred{Cancer}(\const{A})=\true, \pred{Cancer}(\const{B})=\true$ is a model. Assuming $\Phi(\pred{Smokes})=0.2$, $\Phi(\pred{Cancer})=0.8$, $\overline{\Phi}(\pred{Smokes})=0.5$ and $\overline{\Phi}(\pred{Cancer})=1.2$, the weight of this model is $0.2 \cdot 0.5 \cdot 0.8 \cdot 0.8$. This theory has eight other models. The WFOMC can be calculated by summing the weights of all nine models.
\end{example}

\subsection{Converting Inference for SRL Models into WFOMC}
For many SRL models, (lifted) inference can be converted into a WFOMC problem. As an example, consider a Markov logic network (MLN) \citep{Richardson:2006aa} with weighted formulae $(w_1: F_1, \dots, w_k: F_k)$. For every weighted formula $w_i: F_i$ of this MLN, let theory $T$ have a sentence $\pred{Aux}_i \Leftrightarrow F_i$ such that $\pred{Aux}_i$ is a predicate having all LVs appearing in $F_i$. Assuming $\Phi(\pred{Aux}_i)=\exp(w_i)$, and $\Phi$ and $\overline{\Phi}$ are $1$ for the other predicates, the \emph{partition function} of the MLN is equal to $\wfomc(T)$.

\subsection{Calculating the WFOMC of a Theory} \label{rules}
We now describe a set of rules~\rules\ that can be applied to a theory to find its WFOMC efficiently; for more details, readers are directed to \citep{Van:2011}, \citep{Poole:2011} or \cite{PTP}.
We use the following theory $T$ with two clauses and four PRVs ($\pred{S}(\lv{x},\lv{m})$, $\pred{R}(\lv{x},\lv{m})$, $\pred{T}(\lv{x})$ and $\pred{Q}(\lv{x})$) as our running example:
\begin{center}
$\forall \lv{x} \in \pop{x}, \lv{m} \in \pop{m}: \pred{Q}(\lv{x}) \vee \pred{R}(\lv{x}, \lv{m}) \vee \pred{S}(\lv{x}, \lv{m})$ \quad \quad $\forall \lv{x} \in \pop{x}, \lv{m} \in \pop{m}: \pred{S}(\lv{x}, \lv{m}) \vee \pred{T}(\lv{x})$
\end{center}

\paragraph{Lifted Decomposition}
Assume we ground $\lv{x}$ in $T$. Then the clauses mentioning an arbitrary $\ind{X}_i \in \pop{x}$ are $\forall \lv{m} \in \pop{m}: \pred{Q}(\ind{X}_i) \vee \pred{R}(\ind{X}_i, \lv{m}) \vee \pred{S}(\ind{X}_i, \lv{m})$ and $\forall \lv{m} \in \pop{m}: \pred{S}(\ind{X}_i, \lv{m}) \vee \pred{T}(\ind{X}_i)$.
These clauses are totally disconnected from clauses mentioning $\ind{X}_j \in \pop{x}$ ($j\neq i$), and are the same up to renaming $\ind{X}_i$ to $\ind{X}_j$. Given the exchangeability of the individuals, we can calculate the WFOMC of only the clauses mentioning $\ind{X}_i$ and raise the result to the power of the number of connected components ($|\pop{x}|$). Assuming $T_1$ is the theory that results from substituting $\lv{x}$ with $\ind{X}_i$, $\wfomc(T)=\wfomc(T_1)^{|\pop{x}|}$. 

\paragraph{Case-Analysis}
The WFOMC of $T_1$ can be computed by a case-analysis over different assignments of values to a ground PRV, e.g., $\pred{Q}(\ind{X}_i)$. 
Let $T_2$ and $T_3$ represent $T_1\wedge \pred{Q}(\ind{X}_i)$ and $T_1 \wedge \neg \pred{Q}(\ind{X}_i)$ respectively. Then, $\wfomc(T_1) = \wfomc(T_2)+\wfomc(T_3)$. We follow the process for $T_3$ (the process for $T_2$ will be similar) having clauses  $\neg \pred{Q}(\ind{X}_i)$, $\forall \lv{m} \in \pop{m}: \pred{Q}(\ind{X}_i) \vee \pred{R}(\ind{X}_i, \lv{m}) \vee \pred{S}(\ind{X}_i, \lv{m})$ and $\forall \lv{m} \in \pop{m}: \pred{S}(\ind{X}_i, \lv{m}) \vee \pred{T}(\ind{X}_i)$.

\paragraph{Unit Propagation}
When a clause in the theory has only one literal, we can propagate the effect of this clause through the theory and remove it\footnote{Note that unit propagation may remove clauses and random variables from the theory. To account for them, \emph{smoothing} multiplies the WFOMC by $2^{\#rv}$, where $\#rv$ represents the number of removed variables.}. In $T_3$, $\neg \pred{Q}(\ind{X}_i)$ is a unit clause. Having this unit clause, we can simplify the second clause and get the theory $T_4$ having clauses $\forall \lv{m} \in \pop{m}: \pred{R}(\ind{X}_i, \lv{m}) \vee \pred{S}(\ind{X}_i, \lv{m})$ and $\forall \lv{m} \in \pop{m}: \pred{S}(\ind{X}_i, \lv{m}) \vee \pred{T}(\ind{X}_i)$.

\paragraph{Lifted Case-Analysis}
Case-analysis can be done for PRVs having one logical variable in a lifted way. Consider the $\pred{S}(\ind{X}_i, \lv{m})$ in $T_4$. Due to the exchangeability of the individuals, we do not have to consider all possible assignments to all ground PRVs of $\pred{S}(\ind{X}_i, \lv{m})$, but only the ones where the number of individuals $\ind{M} \in \pop{m}$ for which $\pred{S}(\ind{X}_i, \ind{M})$ is $\true$ (or equivalently $\false$) is different. This means considering $|\pop{m}|+1$ cases suffice, corresponding to $\pred{S}(\ind{X}_i, \ind{M})$ being $\true$ for exactly $j=0,\ldots,|\pop{m}|$ individuals. Note that we must multiply by $\binom{|\pop{m}|}{j}$ to account for the number of ways one can select $j$ out of $|\pop{m}|$ individuals. 
Let $T_{4j}$ represent $T_4$ with two more clauses: $\forall \lv{m} \in \pop{m_T}: \pred{S}(\ind{X}_i, \lv{m})$ and $\forall \lv{m} \in \pop{m_F}: \neg \pred{S}(\ind{X}_i, \lv{m})$, where $\pop{m_T}$ represents the $j$ individuals in $\pop{m}$ for which $\pred{S}(\ind{X}_i, \ind{M})$ is $\true$, and $\pop{m_F}$ represents the other $|\pop{m}|-j$ individuals. Then $\wfomc(T_4)=\sum_{j=0}^{\pop{m}}\binom{\pop{m}}{j}\wfomc(T_{4j})$.

\paragraph{Shattering}
In $T_{4j}$, the individuals in $\pop{m}$ are no longer exchangeable: we know different things about those in $\pop{m_T}$ and those in $\pop{m_F}$. We need to shatter every clause having individuals coming from $\pop{m}$ to make the theory exchangeable. To do so, the clause $\forall \lv{m} \in \pop{m}: \pred{R}(\ind{X}_i, \lv{m}) \vee \pred{S}(\ind{X}_i, \lv{m})$ in $T_{4j}$ must be shattered to $\forall \lv{m} \in \pop{m_T}: \pred{R}(\ind{X}_i, \lv{m}) \vee \pred{S}(\ind{X}_i, \lv{m})$ and $\forall \lv{m} \in \pop{m_F}: \pred{R}(\ind{X}_i, \lv{m}) \vee \pred{S}(\ind{X}_i, \lv{m})$ (and similarly for the other formulae).
The shattered theory $T_{5j}$ after unit propagation will have clauses $\forall \lv{m} \in \pop{m_F}: \pred{R}(\ind{X}_i, \lv{m})$ and $\forall \lv{m} \in \pop{m_F}: \pred{T}(\ind{X}_i)$.

\paragraph{Decomposition, Caching, and Grounding}
In $T_{5j}$, the two clauses have different PRVs, i.e., they are disconnected. In such cases, we apply decomposition, i.e., find the WFOMC of each connected component separately and return the product. The WFOMC of the theory can be found by continuing to apply the above rules.
In all the above steps, after finding the WFOMC of each (sub-)theory, we store the results in a cache so we can reuse them if the same WFOMC is required again. By following these steps, one can find the WFOMC of many theories in polynomial time. However, if we reach a point where none of the above rules are applicable, we ground one of the populations which makes the process exponential in the number of individuals.

\subsection{Domain-Liftability}
\begin{definition}
A theory is \textbf{domain-liftable} \citep{gvdb2011completeness} if calculating its WFOMC is polynomial in $|\pop{x_1}|, |\pop{x_2}|, \dots, |\pop{x_k}|$ where $\lv{x}_1, \lv{x}_2, \dots, \lv{x}_k$ represent the LVs in the theory. A class $C$ of theories is domain-liftable if $\forall T \in C$, $T$ is domain-liftable.
\end{definition}
So far, two classes of domain-liftable theories have been recognized: $\FOtwo$ \citep{gvdb2011completeness,van2014skolemization} and \emph{recursively unary} \citep{Poole:2011,LRC2CPP}. 
\begin{definition}
A theory is in $\FOtwo$ if all its clauses have up to two LVs.
\end{definition}
\begin{definition}
Let $T$ be a theory. $T$ is \emph{recursively unary (RU)} if for every theory $T'$ resulting from applying rules in \rules\ except lifted case-analysis to $T$ until no more rules apply, there exists some unary PRV in $T'$ and a generic case of lifted case-analysis on this unary PRV is \emph{RU}. 
\end{definition}

\begin{definition}
Let $C$ be a domain-liftable class of theories. We define $C$ to be \emph{linear} if for any given theory $T$, determining whether $T \in C$ (i.e. membership checking) can be done in time linear in the size of $T$, and to be \emph{domain size independent} if determining whether $T \in C$ is independent of the size of the domains in $T$. Note that a linear class is domain size independent.
\end{definition}
Given the above definitions, $\FOtwo$ is linear. Membership checking can be done for it by a single pass through the theory, counting the number of LVs of each sentence. $RU$ is not linear as its membership checking may be exponential in the size of theory, but it is domain size independent as none of the operations it applies to the input theory depend on the domain sizes. $\FOtwo$ offers faster membership checking than \emph{RU}, but as we will show later, \emph{RU} subsumes $\FOtwo$. This gives rise to a trade-off between fast membership checking and modelling power for, e.g., (lifted) learning purposes.

\section{The Domain Recursion Rule}
\citet{gvdb2011completeness} considered another rule called \emph{domain recursion} in the set of rules for calculating the WFOMC of a theory. The intuition behind domain recursion is that it modifies a domain $\pop{x}$ by making one element explicit: $\pop{x} = \pop{x'} \cup \{\ind{A}\}$ with $\ind{A} \not\in \pop{x'}$. Then, by applying standard rules in \rules\ on this modified theory, the problem is reduced to a WFOMC problem on the original theory, but on a smaller domain $\pop{x'}$. This lets us compute WFOMC using dynamic programming. We refer to \rules\ extended with the domain recursion rule as \rulesplus.
\begin{example} \label{fxy_then_fyx}
Suppose we have a theory whose only clause is $\forall \lv{x}, \lv{y} \in \pop{p}: \neg \pred{Friend}(\lv{x}, \lv{y}) \vee \pred{Friend}(\lv{y}, \lv{x})$, stating if $\lv{x}$ is friends with $\lv{y}$, $\lv{y}$ is also friends with $\lv{x}$. One way to calculate the WFOMC of this theory is by grounding only one individual in $\pop{p}$ and then using \rules. Let $\ind{A}$ be an individual in $\pop{p}$ and let $\pop{p'}=\pop{p}-\{\ind{A}\}$. We can (using domain recursion) rewrite the theory as:
$\forall \lv{x} \in \pop{p'}: \neg \pred{Friend}(\lv{x}, \ind{A}) \vee \pred{Friend}(\ind{A}, \lv{x})$, $\forall \lv{y} \in \pop{p'}: \neg \pred{Friend}(\ind{A}, \lv{y}) \vee \pred{Friend}(\lv{y}, \ind{A})$, and
$\forall \lv{x}, \lv{y} \in \pop{p'}: \neg \pred{Friend}(\lv{x}, \lv{y}) \vee \pred{Friend}(\lv{y}, \lv{x})$.
Lifted case-analysis on $\pred{Friend}(p', \ind{A})$ and $\pred{Friend}(\ind{A}, p')$, shattering and unit propagation give $\forall \lv{x}, \lv{y} \in \pop{p'}: \neg \pred{Friend}(\lv{x}, \lv{y}) \vee \pred{Friend}(\lv{y}, \lv{x})$.
This theory is equivalent to our initial theory, with the only difference being that the population of people has decreased by one. By keeping a cache of the values of each sub-theory, one can verify that this process finds the WFOMC of the above theory in polynomial-time.
\end{example}

Note that the theory in Example~\ref{fxy_then_fyx} is in $\FOtwo$ and as proved in \cite{taghipour2013completeness}, its WFOMC can be computed without using the domain recursion rule\footnote{This can be done by realizing that the theory is disconnected in the grounding for every pair $(\ind{A},\ind{B})$ of individuals and applying the lifted case-analysis.}. This proof has caused the domain recursion rule to be forgotten, or even unknown in lifted inference community. In the next section, we revive this rule and identify a class of theories that are only domain-liftable when using the domain recursion rule.

\section{Domain Recursion Makes More Theories Domain-Liftable} \label{three-theories}
\textbf{S4 Clause:} \citet{beame2015symmetric} identified a clause (S4) with four binary PRVs having the same predicate and proved that even though the rules \rules\ in Section~\ref{rules} cannot calculate the WFOMC of that clause, there is a polynomial-time algorithm for finding its WFOMC. They concluded that this set of rules \rules\ for finding the WFOMC of theories does not suffice, asking for new rules to compute their theory. We prove that adding domain recursion to the set achieves this goal. 

\begin{proposition} \label{S4-prop}
The theory consisting of the S4 clause $\forall \lv{x}_1, \lv{x}_2 \in \pop{x}, \lv{y}_1, \lv{y}_2 \in \pop{y}: \pred{S}(\lv{x}_1, \lv{y}_1) \vee \neg \pred{S}(\lv{x}_2, \lv{y}_1) \vee \pred{S}(\lv{x}_2, \lv{y}_2) \vee \neg \pred{S}(\lv{x}_1, \lv{y}_2)$ is domain-liftable using \rulesplus.
\end{proposition}

\textbf{Symmetric Transitivity:} Domain-liftable calculation of WFOMC for the transitivity formula is a long-lasting open problem. Symmetric-transitivity is easier as the number of its models corresponds to the Bell number, but solving it using general-purpose rules has been an open problem. Consider clauses $\forall \lv{x}, \lv{y}, \lv{z} \in \pop{p}: \neg \pred{F}(\lv{x}, \lv{y}) \vee \neg \pred{F}(\lv{y}, \lv{z}) \vee \pred{F}(\lv{x}, \lv{z})$ and $\forall \lv{x}, \lv{y} \in \pop{p}: \neg \pred{F}(\lv{x}, \lv{y}) \vee \pred{F}(\lv{y}, \lv{x})$ defining a symmetric-transitivity relation. For example, $\pop{p}$ may indicate the population of people and $\pred{F}$ may indicate friendship.
\begin{proposition} \label{sym-trans-prop}
The symmetric-transitivity theory is domain-liftable using \rulesplus.
\end{proposition}

\textbf{Birthday Paradox:} The birthday paradox problem \cite{birthday-paradox} concerns finding the probability that in a set of $n$ randomly chosen people, a pair of them have the same birthday. A first-order encoding of this problem requires WFOMC for a theory with clauses
$\forall \lv{p} \in \pop{p}, \exists \lv{d} \in \pop{d}: \pred{Born}(\lv{p}, \lv{d})$,
$\forall \lv{p} \in \pop{p}, \lv{d}_1, \lv{d}_2 \in \pop{d}: \neg \pred{Born}(\lv{p}, \lv{d}_1) \vee \neg \pred{Born}(\lv{p}, \lv{d}_2)$, and
$\forall \lv{p}_1, \lv{p}_2 \in \pop{p}, \lv{d} \in \pop{d}: \neg \pred{Born}(\lv{p}_1, \lv{d}) \vee \neg \pred{Born}(\lv{p}_2, \lv{d})$, 
where $\pop{p}$ and $\pop{d}$ represent the population of people and days.
The first two clauses impose the condition that every person is born in exactly one day, and the third clause imposes the "no two people are born in the same day" query.
\begin{proposition} \label{birth-para-prop}
The birthday-paradox theory is domain-liftable using \rulesplus.
\end{proposition}

\section{$\mathbf{\StwoFOtwo}$ and $\mathbf{\StwoRU}$: New Domain-Liftable Classes}
\begin{definition} \label{s2fo2-def}
Let $\alpha(\pred{S})$ represent a clausal theory using a single binary predicate $\pred{S}$ such that each clause has exactly two different literals of $\pred{S}$, let $\alpha = \alpha(\pred{S}_1) \wedge \alpha(\pred{S}_2) \wedge \dots \wedge \alpha(\pred{S}_n)$ where $\pred{S}_i$s are different binary predicates, and let $\beta$ represent a theory where all clauses $c \in \beta$ contain at most one $\pred{S}_i$ literal, and the clauses $c \in \beta$ that contain an $\pred{S}_i$ literal contain no other literals with more than one LV. Then, $\StwoFOtwo$ and $\StwoRU$ are the classes of theories of the form $\alpha \wedge \beta$ where $\beta \in \FOtwo$ and $\beta \in RU$ respectively.
\end{definition}

\begin{theorem} \label{liftable-class}
$\StwoFOtwo$ and $\StwoRU$ are domain-liftable using \rulesplus.
\end{theorem}

It can be easily verified that $\StwoFOtwo$ is a linear and $\StwoRU$ is a domain size independent class.

\begin{example} \label{volunteers-jobs}
Suppose we have a set $\pop{j}$ of jobs and a set $\pop{v}$ of volunteers. Every volunteer must be assigned to at most one job, and every job requires no more than one person. If the job involves working with gas, the assigned volunteer must be a non-smoker. And we know that smokers are most probably friends with each other. Then we will have the following first-order theory: 
\begin{center}
$\forall \lv{v}_1, \lv{v}_2 \in \pop{v}, \lv{j} \in \pop{j}: \neg \pred{Assigned}(\lv{v}_1, \lv{j}) \vee \neg \pred{Assigned}(\lv{v}_2, \lv{j})$ \\
$\forall \lv{v} \in \pop{v}, \lv{j}_1, \lv{j}_2 \in \pop{j}: \neg \pred{Assigned}(\lv{v}, \lv{j}_1) \vee \neg \pred{Assigned}(\lv{v}, \lv{j}_2)$ \\
$\forall \lv{v} \in \pop{v}, \lv{j} \in \pop{j}: \pred{InvolvesGas}(\lv{j}) \wedge \pred{Assigned}(\lv{v}, \lv{j}) \Rightarrow \neg \pred{Smokes}(\lv{v})$ \\
$\forall \lv{v}_1, \lv{v}_2 \in \pop{v}: \pred{Aux}(\lv{v}_1, \lv{v}_2) \Leftrightarrow (\pred{Smokes}(\lv{v}_1) \wedge \pred{Friends}(\lv{v}_1, \lv{v}_2) \Rightarrow \pred{Smokes}(\lv{v}_2))$
\end{center}
$\pred{Aux}(\lv{v}_1,\lv{v}_2)$ is added to capture the probability assigned to the last rule.
This theory is not in $\FOtwo$ and not in $RU$ and is not domain-liftable using \rules. However, the first two clauses are instances of $\alpha{Assigned}$ in Def.~\ref{s2fo2-def}, the third and fourth are in $\FOtwo$ (and also in $RU$), and the third clause which contains $\pred{Assigned}(\lv{v}, \lv{j})$ has no other PRVs with more than one LV. Therefore, this theory is in $\StwoFOtwo$ (and also in $\StwoRU$) and domain-liftable based on Theorem~\ref{liftable-class}.
\end{example}

\begin{proposition} \label{subsets-prop}
$\FOtwo \subset RU$, $\FOtwo \subset \StwoFOtwo$, $\FOtwo \subset \StwoRU$, $RU \subset \StwoRU$, $\StwoFOtwo \subset \StwoRU$.
\end{proposition}

\section{Experiments and Results}
In order to see the effect of using domain recursion in practice, we find the WFOMC of three theories with and without using the domain recursion rule: 1- the theory in Example~\ref{volunteers-jobs}, 2- the S4 clause, and 3- the symmetric-transitivity. We implemented the domain recursion rule in C++ and compiled the codes using the g++ compiler. We compare our results with the WFOMC-v3.0 software\footnote{Available at: https://dtai.cs.kuleuven.be/software/wfomc}. Since this software requires domain-liftable input theories, for the first theory we grounded the jobs, for the second we grounded $\pop{x}$, and for the third we grounded $\pop{p}$. For each of these three theories, assuming $|\pop{x}|=n$ for all LVs $\lv{x}$ in the theory, we varied $n$ and plotted the run-time as a function of $n$. All experiments were done on a 2.8GH core with 4GB RAM under MacOSX. The run-times are reported in seconds. We allowed a maximum of $1000s$ for each run.

\begin{figure}
\begin{center}
\includegraphics[width=\textwidth]{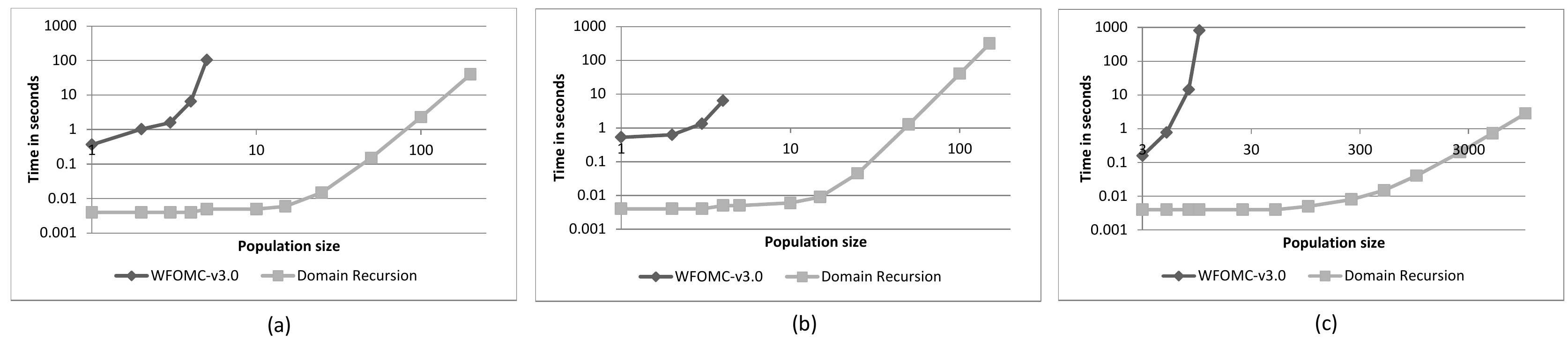}
\end{center}
\caption{Run-times for calculating the WFOMC of (a) the theory in Example~\ref{volunteers-jobs}, (b) the S4 clause, and (c) the symmetric-transitivity, using the WFOMC-v3.0 software (which only uses \rules) and comparing it to the case where we use the domain recursion rule, referred to as \emph{Domain Recursion} in the diagrams.}
\label{results}
\end{figure}

Obtained results can be viewed in Fig.~\ref{results}. These results are consistent with our theory and indicate the clear advantage of using the domain recursion rule in practice. In Fig.~\ref{results}(a), the slope of the diagram for domain recursion is approximately $4$ which indicates the degree of the polynomial for the time complexity. Similar analysis can be done for the results on $S4$ clause and the symmetric-transitivity clause represented in Fig.~\ref{results}(b), (c). The slope of the diagram in these two diagrams is around $5$ and $2$ respectively, indicating that the time complexity for finding the WFOMC of the $S4$ clause and the symmetric-transitivity theories are $n^5$ and $n^2$ respectively, where $n$ shows the size of the population.

\section{Discussion}
We can categorize the theories with respect to the domain recursion rule as: 1- theories proved to be domain-liftable using domain recursion (e.g., S4, symmetric-transitivity, and theories in $\StwoFOtwo$), 2- theories that are domain-liftable using domain recursion, but we have not identified them yet, and 3- theories that are not domain-liftable even when using domain recursion. We leave discovering and characterizing the theories in category 2 and 3 as future work. But here we show that even though the theories in category 3 are not domain-liftable using domain recursion, this rule may still result in exponential speedup for these theories.

Consider the (non-symmetric) transitivity rule:
$\forall \lv{x}, \lv{y}, \lv{z} \in \pop{p}: \neg \pred{Friend}(\lv{x}, \lv{y}) \vee \neg \pred{Friend}(\lv{y}, \lv{z}) \vee \pred{Friend}(\lv{x}, \lv{z})$. 
Since none of the rules in \rules\ apply to the above theory, the existing lifted inference engines ground $\pop{p}$ and calculate the weighted model count (WMC) of the ground theory. By grounding $\pop{p}$, these engines lose great amounts of symmetry. 
Suppose $\pop{p}=\{\const{A},\const{B},\const{C}\}$ and assume we select $\pred{Friend}(\const{A},\const{B})$ and $\pred{Friend}(\const{A},\const{C})$ as the first two random variables for case-analysis. Due to the exchangeability of the individuals, the case where $\pred{Friend}(\const{A},\const{B})$ and $\pred{Friend}(\const{A},\const{C})$ are assigned to $\true$ and $\false$ respectively has the same WMC as the case where they are assigned to $\false$ and $\true$. However, the current engines fail to exploit this symmetry as they consider grounded individuals non-exchangeable. 
By applying domain recursion to the above theory, one can exploit the symmetries of the theory. Suppose $\pop{p'}=\pop{p}-\{\ind{P}\}$. Then we can rewrite the theory as follows:
\begin{center}
$\forall \lv{y}, \lv{z} \in \pop{p'}: \neg \pred{Friend}(\ind{P}, \lv{y}) \vee \neg \pred{Friend}(\lv{y}, \lv{z}) \vee \pred{Friend}(\ind{P}, \lv{z})$ \\
$\forall \lv{x}, \lv{z} \in \pop{p'}: \neg \pred{Friend}(\lv{x}, \ind{P}) \vee \neg \pred{Friend}(\ind{P}, \lv{z}) \vee \pred{Friend}(\lv{x}, \lv{z})$ \\
$\forall \lv{x}, \lv{y} \in \pop{p'}: \neg \pred{Friend}(\lv{x}, \lv{y}) \vee \neg \pred{Friend}(\lv{y}, \ind{P}) \vee \pred{Friend}(\lv{x}, \ind{P})$ \\
$\forall \lv{x}, \lv{y}, \lv{z} \in \pop{p'}: \neg \pred{Friend}(\lv{x}, \lv{y}) \vee \neg \pred{Friend}(\lv{y}, \lv{z}) \vee \pred{Friend}(\lv{x}, \lv{z})$ \\
\end{center} 
By applying lifted case-analysis on $\pred{Friend}(\ind{P}, \lv{y})$, we do not get back the same theory with reduced population and calculating the WFOMC is still exponential. However, we only generate one branch for the case where $\pred{Friend}(\ind{P}, \lv{y})$ is $\true$ only once. This branch covers both the symmetric cases mentioned above. Exploiting these symmetries reduces the time-complexity exponentially. 
This suggests that for any given theory, when the rules in \rules\ are not applicable one may want to try the domain recursion rule before giving up and resorting to grounding a population.

\section{Conclusion}
We identified new classes of domain-liftable theories called $\StwoFOtwo$ and $\StwoRU$ by reviving the domain recursion rule. We also demonstrated how this rule is useful for theories outside these classes. Our work opens up a future research direction for identifying and characterizing larger classes of theories that are domain-liftable using domain recursion. It also helps us get closer to finding a dichotomy between the theories that are domain-liftable and those that are not, similar to the dichotomy result of \citet{dalvi2007efficient} for query answering in probabilistic databases.

It has been shown \citep{LRC2CPP,Kazemi:2016} that compiling the WFOMC rules into low-level programs (e.g., C++ programs) offers (approx.) 175x speedup compared to other approaches. While compiling the previously known rules to low-level programs was straightforward, compiling the domain recursion rule to low-level programs without using recursion might be tricky as it relies on the population size of the logical variables. A future research direction would be finding if the domain recursion rule can be efficiently compiled into low-level programs, and measuring the amount of speedup it offers.

\section{Proofs of the Theorems, Propositions, and Lemmas}
\subsection{Proof of Proposition~\ref{S4-prop}}
\begin{proof}
Let $\pop{x'} = \pop{x} - \{\const{N}\}$. Applying \emph{domain recursion} on $\pop{x}$ (choosing $\pop{y}$ is analogous) gives the following \emph{shattered} theory on the reduced domain $\pop{x'}$.
\begin{align}
  \forall \lv{x_1}, \lv{x_2} \in \pop{x'},~ \lv{y_1}, \lv{y_2} \in \pop{y}:&~ \pred{S}(\lv{x_1},\lv{y_1}) \lor \neg \pred{S}(\lv{x_1},\lv{y_2}) \lor \neg \pred{S}(\lv{x_2},\lv{y_1}) \lor \pred{S}(\lv{x_2},\lv{y_2}) \\
  \forall \lv{x_2} \in \pop{x'},~ \lv{y_1}, \lv{y_2} \in \pop{y}:&~ \pred{S}(\const{N},\lv{y_1}) \lor \neg \pred{S}(\const{N},\lv{y_2}) \lor \neg \pred{S}(\lv{x_2},\lv{y_1}) \lor \pred{S}(\lv{x_2},\lv{y_2}) \\
  \forall \lv{x_1} \in \pop{x'},~ \lv{y_1}, \lv{y_2} \in \pop{y}:&~ \pred{S}(\lv{x_1},\lv{y_1}) \lor \neg \pred{S}(\lv{x_1},\lv{y_2}) \lor \neg \pred{S}(\const{N},\lv{y_1}) \lor \pred{S}(\const{N},\lv{y_2}) \\
  \forall \lv{y_1}, \lv{y_2} \in \pop{y}:&~ \pred{S}(\const{N},\lv{y_1}) \lor \neg \pred{S}(\const{N},\lv{y_2}) \lor \neg \pred{S}(\const{N},\lv{y_1}) \lor \pred{S}(\const{N},\lv{y_2}) 
\end{align}

We now reach to the standard rules \rules\ to simplify the output of domain recursion.
The last clause is a \emph{tautology} and can be dropped.
The theory contains a unary PRV, namely $\pred{S}(\const{N},\lv{y})$, which is a candidate for \emph{lifted case-analysis}. Let $\pop{y_T} \subseteq \pop{y}$ be the individuals of $\pop{y}$ for which $\pred{S}(\const{N},\lv{y})$ is satisfied, and let $\pop{y_F} = \pop{y} \setminus \pop{y_T}$ be its complement in $\pop{y}$. This gives
\begin{align}
  \forall \lv{x_1}, \lv{x_2} \in \pop{x'},~ \lv{y_1}, \lv{y_2} \in \pop{y}:&~ \pred{S}(\lv{x_1},\lv{y_1}) \lor \neg \pred{S}(\lv{x_1},\lv{y_2}) \lor \neg \pred{S}(\lv{x_2},\lv{y_1}) \lor \pred{S}(\lv{x_2},\lv{y_2}) \\
  \forall \lv{x_2} \in \pop{x'},~ \lv{y_1}, \lv{y_2} \in \pop{y}:&~ \pred{S}(\const{N},\lv{y_1}) \lor \neg \pred{S}(\const{N},\lv{y_2}) \lor \neg \pred{S}(\lv{x_2},\lv{y_1}) \lor \pred{S}(\lv{x_2},\lv{y_2}) \\
  \forall \lv{x_1} \in \pop{x'},~ \lv{y_1}, \lv{y_2} \in \pop{y}:&~ \pred{S}(\lv{x_1},\lv{y_1}) \lor \neg \pred{S}(\lv{x_1},\lv{y_2}) \lor \neg \pred{S}(\const{N},\lv{y_1}) \lor \pred{S}(\const{N},\lv{y_2}) \\
  \forall \lv{y} \in \pop{y_T}:&~ \pred{S}(\const{N},\lv{y})\\
  \forall \lv{y} \in \pop{y_F}:&~ \neg \pred{S}(\const{N},\lv{y})
\end{align}

\emph{Unit propagation} creates two independent theories: one containing the $\pred{S}(\const{N},\lv{y})$ atoms, which is trivially liftable, and one containing the other atoms, namely
\begin{align}  
  \forall \lv{x_1}, \lv{x_2} \in \pop{x'},~ \lv{y_1}, \lv{y_2} \in \pop{y}:&~ \pred{S}(\lv{x_1},\lv{y_1}) \lor \neg \pred{S}(\lv{x_1},\lv{y_2}) \lor \neg \pred{S}(\lv{x_2},\lv{y_1}) \lor \pred{S}(\lv{x_2},\lv{y_2}) \\
  \forall \lv{x_2} \in \pop{x'},~ \lv{y_1} \in \pop{y_T}, \lv{y_2} \in \pop{y_F}:&~ \neg \pred{S}(\lv{x_2},\lv{y_1}) \lor \pred{S}(\lv{x_2},\lv{y_2}) \\
  \forall \lv{x_1} \in \pop{x'},~ \lv{y_1} \in \pop{y_T}, \lv{y_2} \in \pop{y_F}:&~ \pred{S}(\lv{x_1},\lv{y_1}) \lor \neg \pred{S}(\lv{x_1},\lv{y_2})
\end{align}
The last two clauses are equivalent, hence, we have
\begin{align}  
  \forall \lv{x_1}, \lv{x_2} \in \pop{x'},~ \lv{y_1}, \lv{y_2} \in \pop{y}:&~ \pred{S}(\lv{x_1},\lv{y_1}) \lor \neg \pred{S}(\lv{x_1},\lv{y_2}) \lor \neg \pred{S}(\lv{x_2},\lv{y_1}) \lor \pred{S}(\lv{x_2},\lv{y_2}) \\
  \forall \lv{x} \in \pop{x'},~ \lv{y_1} \in \pop{y_F}, \lv{y_2} \in \pop{y_T}:&~ \neg \pred{S}(\lv{x},\lv{y_1}) \lor \pred{S}(\lv{x},\lv{y_2})
\end{align}
After \emph{shattering}, we get four copies of the first clause:
\begin{align}  
  \forall \lv{x_1}, \lv{x_2} \in \pop{x'},~ \lv{y_1} \in \pop{y_T}, \lv{y_2} \in \pop{y_T}:&~ \pred{S}(\lv{x_1},\lv{y_1}) \lor \neg \pred{S}(\lv{x_1},\lv{y_2}) \lor \neg \pred{S}(\lv{x_2},\lv{y_1}) \lor \pred{S}(x_2,y_2) \\
  \forall \lv{x_1}, \lv{x_2} \in \pop{x'},~ \lv{y_1} \in \pop{y_T}, \lv{y_2} \in \pop{y_F}:&~ \pred{S}(\lv{x_1},\lv{y_1}) \lor \neg \pred{S}(\lv{x_1},\lv{y_2}) \lor \neg \pred{S}(\lv{x_2},\lv{y_1}) \lor \pred{S}(\lv{x_2},\lv{y_2}) \\
  \forall \lv{x_1}, \lv{x_2} \in \pop{x'},~ \lv{y_1} \in \pop{y_F}, \lv{y_2} \in \pop{y_T}:&~ \pred{S}(\lv{x_1},\lv{y_1}) \lor \neg \pred{S}(\lv{x_1},\lv{y_2}) \lor \neg \pred{S}(\lv{x_2},\lv{y_1}) \lor \pred{S}(\lv{x_2},\lv{y_2}) \\
  \forall \lv{x_1}, \lv{x_2} \in \pop{x'},~ \lv{y_1} \in \pop{y_F}, \lv{y_2} \in \pop{y_F}:&~ \pred{S}(\lv{x_1},\lv{y_1}) \lor \neg \pred{S}(\lv{x_1},\lv{y_2}) \lor \neg \pred{S}(\lv{x_2},\lv{y_1}) \lor \pred{S}(\lv{x_2},\lv{y_2}) \\
  \forall \lv{x} \in \pop{x'},~ \lv{y_1} \in \pop{y_F}, \lv{y_2} \in \pop{y_T}:&~ \neg \pred{S}(\lv{x},\lv{y_1}) \lor \pred{S}(\lv{x},\lv{y_2})
\end{align}
The second and third clauses are subsumed by the last clause, and can be removed:
\begin{align}
  \forall \lv{x_1}, \lv{x_2} \in \pop{x'},~ \lv{y_1} \in \pop{y_T}, \lv{y_2} \in \pop{y_T}:&~ \pred{S}(\lv{x_1},\lv{y_1}) \lor \neg \pred{S}(\lv{x_1},\lv{y_2}) \lor \neg \pred{S}(\lv{x_2},\lv{y_1}) \lor \pred{S}(\lv{x_2},\lv{y_2}) \label{removed1}\\
  \forall \lv{x_1}, \lv{x_2} \in \pop{x'},~ \lv{y_1} \in \pop{y_F}, \lv{y_2} \in \pop{y_F}:&~ \pred{S}(\lv{x_1},\lv{y_1}) \lor \neg \pred{S}(\lv{x_1},\lv{y_2}) \lor \neg \pred{S}(\lv{x_2},\lv{y_1}) \lor \pred{S}(\lv{x_2},\lv{y_2}) \label{removed2}\\
  \forall \lv{x} \in \pop{x'},~ \lv{y_1} \in \pop{y_F}, \lv{y_2} \in \pop{y_T}:&~ \neg \pred{S}(\lv{x},\lv{y_1}) \lor \pred{S}(\lv{x},\lv{y_2})
\end{align}
Let us now consider the last clause, and ignore the first two for the time being. The last clause is actually in FO$^2$, and the \emph{Skolemization}-rewriting of reused FO$^2$ variables \cite{van2014skolemization} can be applied, for example to $\lv{y_2}$ in its second PRV. The last clause is thus replaced by
\begin{align}  
  \forall \lv{x} \in \pop{x'},~ \lv{y} \in \pop{y_F}:&~ \neg \pred{S}(\lv{x},\lv{y}) \lor \neg \pred{A}(\lv{x})\\
  \forall \lv{x} \in \pop{x'},~ \lv{y} \in \pop{y_T}:&~ \pred{S}(\lv{x},\lv{y}) \lor \pred{A}(\lv{x}) \\
  \forall \lv{x} \in \pop{x'}:&~ \pred{A}(\lv{x}) \lor \pred{B}(\lv{x}) \\
  \forall \lv{x} \in \pop{x'},~ \lv{y} \in \pop{y_T}:&~  \pred{S}(\lv{x},\lv{y})\lor \pred{B}(\lv{x})
\end{align}
Next, we perform \emph{lifted case-analysis} on $\pred{A}(\lv{x'})$. Let $\pop{\alpha} \subseteq \pop{x'}$ be the individuals in $\pop{x'}$ for which $\pred{A}(\lv{x'})$ is satisfied, and let $\pop{\bar{\alpha}} = \pop{x'} \setminus \pop{\alpha}$ be its complement in $\pop{x'}$:
\begin{align}  
  \forall \lv{x} \in \pop{x'},~ \lv{y} \in \pop{y_F}:&~ \neg \pred{S}(\lv{x},\lv{y}) \lor \neg \pred{A}(\lv{x})\\
  \forall \lv{x} \in \pop{x'},~ \lv{y} \in \pop{y_T}:&~ \pred{S}(\lv{x},\lv{y}) \lor \pred{A}(\lv{x}) \\
  \forall \lv{x} \in \pop{x'}:&~ \pred{A}(\lv{x}) \lor \pred{B}(\lv{x}) \\
  \forall \lv{x} \in \pop{x'},~ \lv{y} \in \pop{y_T}:&~  \pred{S}(\lv{x},\lv{y})\lor \pred{B}(\lv{x})\\
  \forall \lv{x} \in \pop{\alpha}:&~ \pred{A}(\lv{x})\\
  \forall \lv{x} \in \pop{\bar{\alpha}}:&~ \neg \pred{A}(\lv{x})
\end{align}
\emph{Unit propagation} gives two independent theories: a theory containing the predicate $\pred{A}$, which is trivially liftable, and the theory
\begin{align}  
  \forall \lv{x} \in \pop{\alpha},~ \lv{y} \in \pop{y_F}:&~ \neg \pred{S}(\lv{x},\lv{y})\\
  \forall \lv{x} \in \pop{\bar{\alpha}},~ \lv{y} \in \pop{y_T}:&~ \pred{S}(\lv{x},\lv{y}) \\
  \forall \lv{x} \in \pop{\bar{\alpha}}:&~ \pred{B}(\lv{x}) \\
  \forall \lv{x} \in \pop{\alpha},~ \lv{y} \in \pop{y_T}:&~  \pred{S}(\lv{x},\lv{y}) \lor \pred{B}(\lv{x})
\end{align}
Next, we perform \emph{atom counting} on $\pred{B}(\lv{x'})$. Let $\pop{\beta} \subseteq \pop{\alpha}$ be the individuals of $\pop{\alpha}$ for which $\pred{B}(\lv{x'})$ is satisfied, and let $\pop{\bar{\beta}} = \pop{\alpha} \setminus \pop{\beta}$ be its complement in $\pop{\alpha}$. In other words, the original domain $\pop{x}$ is now split up into four parts: $\pop{\bar{x}}=\{\const{N}\}$, $\pop{\bar{\alpha}}$, $\pop{\beta}$, and $\pop{\bar{\beta}}$. This gives the theory
\begin{align}  
  \forall \lv{x} \in \pop{\alpha},~ \lv{y} \in \pop{y_F}:&~ \neg \pred{S}(\lv{x},\lv{y})\\
  \forall \lv{x} \in \pop{\bar{\alpha}},~ \lv{y} \in \pop{y_T}:&~ \pred{S}(\lv{x},\lv{y}) \\
  \forall \lv{x} \in \pop{\bar{\alpha}}:&~ \pred{B}(\lv{x}) \\
  \forall \lv{x} \in \pop{\alpha},~ \lv{y} \in \pop{y_T}:&~  \pred{S}(\lv{x},\lv{y}) \lor \pred{B}(\lv{x})\\
  \forall \lv{x} \in \pop{\beta}:&~ \pred{B}(\lv{x})\\
  \forall \lv{x} \in \pop{\bar{\beta}}:&~ \neg \pred{B}(\lv{x})
\end{align}
\emph{Unit propagation} gives two independent theories: a theory containing the predicate $\pred{B}$, which is trivially liftable, and the theory
\begin{align}  
  \forall \lv{x} \in \pop{\alpha},~ \lv{y} \in \pop{y_F}:&~ \neg \pred{S}(\lv{x},\lv{y}) \label{prop01}\\
  \forall \lv{x} \in \pop{\bar{\alpha}},~ \lv{y} \in \pop{y_T}:&~ \pred{S}(\lv{x},\lv{y}) \label{prop02} \\
  \forall \lv{x} \in \pop{\bar{\beta}},~ \lv{y} \in \pop{y_T}:&~  \pred{S}(\lv{x},\lv{y})  \label{prop03}
\end{align}
We now reintroduce the first removed clause. Clause \ref{removed1} has nine copies after \emph{shattering}:
\begin{align}  
  \forall \lv{x_1} \in \pop{\bar{\alpha}}, \lv{x_2} \in \pop{\bar{\alpha}},~ \lv{y_1}, \lv{y_2} \in \pop{y_T}:&~ \pred{S}(\lv{x_1},\lv{y_1}) \lor \neg \pred{S}(\lv{x_1},\lv{y_2}) \lor \neg \pred{S}(\lv{x_2},\lv{y_1}) \lor \pred{S}(\lv{x_2},\lv{y_2}) \\
  \forall \lv{x_1} \in \pop{\bar{\alpha}}, \lv{x_2} \in \pop{\beta},~ \lv{y_1}, \lv{y_2} \in \pop{y_T}:&~ \pred{S}(\lv{x_1},\lv{y_1}) \lor \neg \pred{S}(\lv{x_1},\lv{y_2}) \lor \neg \pred{S}(\lv{x_2},\lv{y_1}) \lor \pred{S}(\lv{x_2},\lv{y_2}) \\
  \forall \lv{x_1} \in \pop{\bar{\alpha}}, \lv{x_2} \in \pop{\bar{\beta}},~ \lv{y_1}, \lv{y_2} \in \pop{y_T}:&~ \pred{S}(\lv{x_1},\lv{y_1}) \lor \neg \pred{S}(\lv{x_1},\lv{y_2}) \lor \neg \pred{S}(\lv{x_2},\lv{y_1}) \lor \pred{S}(\lv{x_2},\lv{y_2}) \\
  \forall \lv{x_1} \in \pop{\beta}, \lv{x_2} \in \pop{\bar{\alpha}},~ \lv{y_1}, \lv{y_2} \in \pop{y_T}:&~ \pred{S}(\lv{x_1},\lv{y_1}) \lor \neg \pred{S}(\lv{x_1},\lv{y_2}) \lor \neg \pred{S}(\lv{x_2},\lv{y_1}) \lor \pred{S}(\lv{x_2},\lv{y_2}) \\
  \forall \lv{x_1} \in \pop{\beta}, \lv{x_2} \in \pop{\beta},~ \lv{y_1}, \lv{y_2} \in \pop{y_T}:&~ \pred{S}(\lv{x_1},\lv{y_1}) \lor \neg \pred{S}(\lv{x_1},\lv{y_2}) \lor \neg \pred{S}(\lv{x_2},\lv{y_1}) \lor \pred{S}(\lv{x_2},\lv{y_2}) \\
  \forall \lv{x_1} \in \pop{\beta}, \lv{x_2} \in \pop{\bar{\beta}},~ \lv{y_1}, \lv{y_2} \in \pop{y_T}:&~ \pred{S}(\lv{x_1},\lv{y_1}) \lor \neg \pred{S}(\lv{x_1},\lv{y_2}) \lor \neg \pred{S}(\lv{x_2},\lv{y_1}) \lor \pred{S}(\lv{x_2},\lv{y_2}) \\
  \forall \lv{x_1} \in \pop{\bar{\beta}}, \lv{x_2} \in \pop{\bar{\alpha}},~ \lv{y_1}, \lv{y_2} \in \pop{y_T}:&~ \pred{S}(\lv{x_1},\lv{y_1}) \lor \neg \pred{S}(\lv{x_1},\lv{y_2}) \lor \neg \pred{S}(\lv{x_2},\lv{y_1}) \lor \pred{S}(\lv{x_2},\lv{y_2}) \\
  \forall \lv{x_1} \in \pop{\bar{\beta}}, \lv{x_2} \in \pop{\beta},~ \lv{y_1}, \lv{y_2} \in \pop{y_T}:&~ \pred{S}(\lv{x_1},\lv{y_1}) \lor \neg \pred{S}(\lv{x_1},\lv{y_2}) \lor \neg \pred{S}(\lv{x_2},\lv{y_1}) \lor \pred{S}(\lv{x_2},\lv{y_2}) \\
  \forall \lv{x_1} \in \pop{\bar{\beta}}, \lv{x_2} \in \pop{\bar{\beta}},~ \lv{y_1}, \lv{y_2} \in \pop{y_T}:&~ \pred{S}(\lv{x_1},\lv{y_1}) \lor \neg \pred{S}(\lv{x_1},\lv{y_2}) \lor \neg \pred{S}(\lv{x_2},\lv{y_1}) \lor \pred{S}(\lv{x_2},\lv{y_2})
\end{align}
\emph{Unit propagation} of clauses \ref{prop02} and \ref{prop03} satisfies any clause that has a positive literal whose $\lv{x}$ domain is $\pop{\bar{\alpha}}$ or $\pop{\bar{\beta}}$. This removes all clauses except for 
\begin{align}  
  \forall \lv{x_1} \in \pop{\beta}, \lv{x_2} \in \pop{\beta},~ \lv{y_1}, \lv{y_2} \in \pop{y_T}:&~ \pred{S}(\lv{x_1},\lv{y_1}) \lor \neg \pred{S}(\lv{x_1},\lv{y_2}) \lor \neg \pred{S}(\lv{x_2},\lv{y_1}) \lor \pred{S}(\lv{x_2},\lv{y_2})
\end{align}
We now reintroduce the second removed clause. Clause \ref{removed2} has four copies after \emph{shattering}:
\begin{align}
  \forall \lv{x_1} \in \pop{\alpha}, \lv{x_2} \in \pop{\alpha},~ \lv{y_1}, \lv{y_2} \in \pop{y_F}:&~ \pred{S}(\lv{x_1},\lv{y_1}) \lor \neg \pred{S}(\lv{x_1},\lv{y_2}) \lor \neg \pred{S}(\lv{x_2},\lv{y_1}) \lor \pred{S}(\lv{x_2},\lv{y_2}) \\
  \forall \lv{x_1} \in \pop{\alpha}, \lv{x_2} \in \pop{\bar{\alpha}},~ \lv{y_1}, \lv{y_2} \in \pop{y_F}:&~ \pred{S}(\lv{x_1},\lv{y_1}) \lor \neg \pred{S}(\lv{x_1},\lv{y_2}) \lor \neg \pred{S}(\lv{x_2},\lv{y_1}) \lor \pred{S}(\lv{x_2},\lv{y_2}) \\
  \forall \lv{x_1} \in \pop{\bar{\alpha}}, \lv{x_2} \in \pop{\alpha},~ \lv{y_1}, \lv{y_2} \in \pop{y_F}:&~ \pred{S}(\lv{x_1},\lv{y_1}) \lor \neg \pred{S}(\lv{x_1},\lv{y_2}) \lor \neg \pred{S}(\lv{x_2},\lv{y_1}) \lor \pred{S}(\lv{x_2},\lv{y_2}) \\
  \forall \lv{x_1} \in \pop{\bar{\alpha}}, \lv{x_2} \in \pop{\bar{\alpha}},~ \lv{y_1}, \lv{y_2} \in \pop{y_F}:&~ \pred{S}(\lv{x_1},\lv{y_1}) \lor \neg \pred{S}(\lv{x_1},\lv{y_2}) \lor \neg \pred{S}(\lv{x_2},\lv{y_1}) \lor \pred{S}(\lv{x_2},\lv{y_2})
\end{align}
\emph{Unit propagation} of clauses \ref{prop01} satisfies any clause that has a negative literal whose $x$ domain is $\alpha$. This removes all clauses except for 
\begin{align}  
  \forall \lv{x_1} \in \pop{\bar{\alpha}}, \lv{x_2} \in \pop{\bar{\alpha}},~ \lv{y_1}, \lv{y_2} \in \pop{y_F}:&~ \pred{S}(\lv{x_1},\lv{y_1}) \lor \neg \pred{S}(\lv{x_1},\lv{y_2}) \lor \neg \pred{S}(\lv{x_2},\lv{y_1}) \lor \pred{S}(\lv{x_2},\lv{y_2})
\end{align}
Putting it all together, we have the theory
\begin{align}  
  \forall \lv{x_1}, \lv{x_2} \in \pop{\beta},~ \lv{y_1}, \lv{y_2} \in \pop{y_T}:&~ \pred{S}(\lv{x_1},\lv{y_1}) \lor \neg \pred{S}(\lv{x_1},\lv{y_2}) \lor \neg \pred{S}(\lv{x_2},\lv{y_1}) \lor \pred{S}(\lv{x_2},\lv{y_2}) \\
  \forall \lv{x_1}, \lv{x_2} \in \pop{\bar{\alpha}},~ \lv{y_1}, \lv{y_2} \in \pop{y_F}:&~ \pred{S}(\lv{x_1},\lv{y_1}) \lor \neg \pred{S}(\lv{x_1},\lv{y_2}) \lor \neg \pred{S}(\lv{x_2},\lv{y_1}) \lor \pred{S}(\lv{x_1},\lv{y_2}) \\
  \forall \lv{x} \in \pop{\alpha},~ \lv{y} \in \pop{y_F}:&~ \neg \pred{S}(\lv{x},\lv{y}) \label{prop1}\\
  \forall \lv{x} \in \pop{\bar{\alpha}},~ \lv{y} \in \pop{y_T}:&~ \pred{S}(\lv{x},\lv{y}) \label{prop2} \\
  \forall \lv{x} \in \pop{\bar{\beta}},~ \lv{y} \in \pop{y_T}:&~  \pred{S}(\lv{x},\lv{y})  \label{prop3}
\end{align}
These five clauses are all \emph{independent}. 
The last three are trivially liftable.
The first two are simply copies of S4 with modified domains $\pop{\beta}$, $\pop{y_T}$, $\pop{\bar{\alpha}}$ and $\pop{y_F}$ instead of $\pop{x}$ and $\pop{y}$. However, we have that $|\pop{\beta}| < |\pop{x}|$, $|\pop{y_T}| \leq |\pop{y}|$, $|\pop{\bar{\alpha}}| < |\pop{x}|$, and $|\pop{y_F}| \leq |\pop{y}|$. The recursion is thus guaranteed to terminate with $\pop{\beta} = \pop{\bar{\alpha}} = \emptyset$. By keeping a cache of WFOMCs for all sizes of $\pop{x}$ and $\pop{y}$, we can compute the WFOMC of S4 in PTIME.
\end{proof}

\subsection{Proof of Proposition~\ref{sym-trans-prop}}
\begin{proof} Symmetric-transitivity has the following two sentences:
\begin{align}
  \forall \lv{x}, \lv{y}, \lv{z} \in \pop{p}:&~ \neg \pred{F}(\lv{x}, \lv{y}) \vee \neg \pred{F}(\lv{y}, \lv{z}) \vee \pred{F}(\lv{x}, \lv{z}) \\
  \forall \lv{x}, \lv{y} \in \pop{p}:&~ \neg \pred{F}(\lv{x}, \lv{y}) \vee \pred{F}(\lv{y}, \lv{x})
\end{align}

Assuming $\pop{q}=\pop{p}-\{N\}$:
\begin{align}
  \forall \lv{y}, \lv{z} \in \pop{q}:&~ \neg \pred{F}(\const{N}, \lv{y}) \vee \neg \pred{F}(\lv{y}, \lv{z}) \vee \pred{F}(\const{N}, \lv{z}) \\
  \forall \lv{x}, \lv{z} \in \pop{q}:&~ \neg \pred{F}(\lv{x}, \const{N}) \vee \neg \pred{F}(\const{N}, \lv{z}) \vee \pred{F}(\lv{x}, \lv{z}) \\
  \forall \lv{x}, \lv{y} \in \pop{q}:&~ \neg \pred{F}(\lv{x}, \lv{y}) \vee \neg \pred{F}(\lv{y}, \const{N}) \vee \pred{F}(\lv{x}, \const{N}) \\
  \forall \lv{x}, \lv{y}, \lv{z} \in \pop{q}:&~ \neg \pred{F}(\lv{x}, \lv{y}) \vee \neg \pred{F}(\lv{y}, \lv{z}) \vee \pred{F}(\lv{x}, \lv{z}) \\
  \forall \lv{y} \in \pop{q}:&~ \neg \pred{F}(\const{N}, \lv{y}) \vee \pred{F}(\lv{y}, \const{N}) \\
  \forall \lv{x} \in \pop{q}:&~ \neg \pred{F}(\lv{x}, \const{N}) \vee \pred{F}(\const{N}, \lv{x}) \\
  \forall \lv{x}, \lv{y} \in \pop{q}:&~ \neg \pred{F}(\lv{x}, \lv{y}) \vee \pred{F}(\lv{y}, \lv{x})
\end{align}
Lifted case-analysis on $\pred{F}(\const{N}, \lv{q})$ assuming $\pop{q_T}$ contains individuals in $\pop{q}$ for which $\pred{F}(\const{N}, \lv{q})$ is true and $\pop{q_F}$ is the other individuals in $\pop{q}$:
\begin{align}
\forall \lv{y}, \lv{z} \in \pop{q}:&~ \neg \pred{F}(\const{N}, \lv{y}) \vee \neg \pred{F}(\lv{y}, \lv{z}) \vee \pred{F}(\const{N}, \lv{z}) \\
\forall \lv{x}, \lv{z} \in \pop{q}:&~ \neg \pred{F}(\lv{x}, \const{N}) \vee \neg \pred{F}(\const{N}, \lv{z}) \vee \pred{F}(\lv{x}, \lv{z}) \\
\forall \lv{x}, \lv{y} \in \pop{q}:&~ \neg \pred{F}(\lv{x}, \lv{y}) \vee \neg \pred{F}(\lv{y}, \const{N}) \vee \pred{F}(\lv{x}, \const{N}) \\
\forall \lv{x}, \lv{y}, \lv{z} \in \pop{q}:&~ \neg \pred{F}(\lv{x}, \lv{y}) \vee \neg \pred{F}(\lv{y}, \lv{z}) \vee \pred{F}(\lv{x}, \lv{z}) \\
\forall \lv{y} \in \pop{q}:&~ \neg \pred{F}(\const{N}, \lv{y}) \vee \pred{F}(\lv{y}, \const{N}) \\
\forall \lv{x} \in \pop{q}:&~ \neg \pred{F}(\lv{x}, \const{N}) \vee \pred{F}(\const{N}, \lv{x}) \\
\forall \lv{x}, \lv{y} \in \pop{q}:&~ \neg \pred{F}(\lv{x}, \lv{y}) \vee \pred{F}(\lv{y}, \lv{x}) \\
\forall \lv{x} \in \pop{q_T}:&~ \pred{F}(\const{N}, \lv{x}) \\
\forall \lv{x} \in \pop{q_F}:&~ \neg \pred{F}(\const{N}, \lv{x})
\end{align}
Unit propagation:
\begin{align}
\forall \lv{y} \in \pop{q_T}, \lv{z} \in \pop{q_F}:&~ \neg \pred{F}(\lv{y}, \lv{z}) \\
\forall \lv{x} \in \pop{q}, \lv{z} \in \pop{q_T}:&~ \neg \pred{F}(\lv{x}, \const{N}) \vee \pred{F}(\lv{x}, \lv{z}) \\
\forall \lv{x}, \lv{y} \in \pop{q}:&~ \neg \pred{F}(\lv{x}, \lv{y}) \vee \neg \pred{F}(\lv{y}, \const{N}) \vee \pred{F}(\lv{x}, \const{N}) \\
\forall \lv{x}, \lv{y}, \lv{z} \in \pop{q}:&~ \neg \pred{F}(\lv{x}, \lv{y}) \vee \neg \pred{F}(\lv{y}, \lv{z}) \vee \pred{F}(\lv{x}, \lv{z}) \\
\forall \lv{y} \in \pop{q_T}:&~ \pred{F}(\lv{y}, \const{N}) \\
\forall \lv{x} \in \pop{q_F}:&~ \neg \pred{F}(\lv{x}, \const{N}) \\
\forall \lv{x}, \lv{y} \in \pop{q}:&~ \neg \pred{F}(\lv{x}, \lv{y}) \vee \pred{F}(\lv{y}, \lv{x})
\end{align}
Shattering:
\begin{align}
\forall \lv{y} \in \pop{q_T}, \lv{z} \in \pop{q_F}:&~ \neg \pred{F}(\lv{y}, \lv{z}) \\
\forall \lv{x} \in \pop{q_T}, \lv{z} \in \pop{q_T}:&~ \neg \pred{F}(\lv{x}, \const{N}) \vee \pred{F}(\lv{x}, \lv{z}) \\
\forall \lv{x} \in \pop{q_F}, \lv{z} \in \pop{q_T}:&~ \neg \pred{F}(\lv{x}, \const{N}) \vee \pred{F}(\lv{x}, \lv{z}) \\
\forall \lv{x} \in \pop{q_T}, \lv{y} \in \pop{q_T}:&~ \neg \pred{F}(\lv{x}, \lv{y}) \vee \neg \pred{F}(\lv{y}, \const{N}) \vee \pred{F}(\lv{x}, \const{N}) \\
\forall \lv{x} \in \pop{q_T}, \lv{y} \in \pop{q_F}:&~ \neg \pred{F}(\lv{x}, \lv{y}) \vee \neg \pred{F}(\lv{y}, \const{N}) \vee \pred{F}(\lv{x}, \const{N}) \\
\forall \lv{x} \in \pop{q_F}, \lv{y} \in \pop{q_T}:&~ \neg \pred{F}(\lv{x}, \lv{y}) \vee \neg \pred{F}(\lv{y}, \const{N}) \vee \pred{F}(\lv{x}, \const{N}) \\
\forall \lv{x} \in \pop{q_F}, \lv{y} \in \pop{q_F}:&~ \neg \pred{F}(\lv{x}, \lv{y}) \vee \neg \pred{F}(\lv{y}, \const{N}) \vee \pred{F}(\lv{x}, \const{N}) \\
\forall \lv{x} \in \pop{q_T}, \lv{y} \in \pop{q_T}, z \in \pop{q_T}:&~ \neg \pred{F}(\lv{x}, \lv{y}) \vee \neg \pred{F}(\lv{y}, \lv{z}) \vee \pred{F}(\lv{x}, \lv{z}) \\
\forall \lv{x} \in \pop{q_T}, \lv{y} \in \pop{q_T}, z \in \pop{q_F}:&~ \neg \pred{F}(\lv{x}, \lv{y}) \vee \neg \pred{F}(\lv{y}, \lv{z}) \vee \pred{F}(\lv{x}, \lv{z}) \\
\forall \lv{x} \in \pop{q_T}, \lv{y} \in \pop{q_F}, z \in \pop{q_T}:&~ \neg \pred{F}(\lv{x}, \lv{y}) \vee \neg \pred{F}(\lv{y}, \lv{z}) \vee \pred{F}(\lv{x}, \lv{z}) \\
\forall \lv{x} \in \pop{q_T}, \lv{y} \in \pop{q_F}, z \in \pop{q_F}:&~ \neg \pred{F}(\lv{x}, \lv{y}) \vee \neg \pred{F}(\lv{y}, \lv{z}) \vee \pred{F}(\lv{x}, \lv{z}) \\
\forall \lv{x} \in \pop{q_F}, \lv{y} \in \pop{q_T}, z \in \pop{q_T}:&~ \neg \pred{F}(\lv{x}, \lv{y}) \vee \neg \pred{F}(\lv{y}, \lv{z}) \vee \pred{F}(\lv{x}, \lv{z}) \\
\forall \lv{x} \in \pop{q_F}, \lv{y} \in \pop{q_T}, z \in \pop{q_F}:&~ \neg \pred{F}(\lv{x}, \lv{y}) \vee \neg \pred{F}(\lv{y}, \lv{z}) \vee \pred{F}(\lv{x}, \lv{z}) \\
\forall \lv{x} \in \pop{q_F}, \lv{y} \in \pop{q_F}, z \in \pop{q_T}:&~ \neg \pred{F}(\lv{x}, \lv{y}) \vee \neg \pred{F}(\lv{y}, \lv{z}) \vee \pred{F}(\lv{x}, \lv{z}) \\
\forall \lv{x} \in \pop{q_F}, \lv{y} \in \pop{q_F}, z \in \pop{q_F}:&~ \neg \pred{F}(\lv{x}, \lv{y}) \vee \neg \pred{F}(\lv{y}, \lv{z}) \vee \pred{F}(\lv{x}, \lv{z}) \\
\forall \lv{y} \in \pop{q_T}:&~ \pred{F}(\lv{y}, \const{N}) \\
\forall \lv{x} \in \pop{q_F}:&~ \neg \pred{F}(\lv{x}, \const{N}) \\
\forall \lv{x} \in \pop{q_T}, \lv{y} \in \pop{q_T}:&~ \neg \pred{F}(\lv{x}, \lv{y}) \vee \pred{F}(\lv{y}, \lv{x}) \\
\forall \lv{x} \in \pop{q_T}, \lv{y} \in \pop{q_F}:&~ \neg \pred{F}(\lv{x}, \lv{y}) \vee \pred{F}(\lv{y}, \lv{x}) \\
\forall \lv{x} \in \pop{q_F}, \lv{y} \in \pop{q_T}:&~ \neg \pred{F}(\lv{x}, \lv{y}) \vee \pred{F}(\lv{y}, \lv{x}) \\
\forall \lv{x} \in \pop{q_F}, \lv{y} \in \pop{q_F}:&~ \neg \pred{F}(\lv{x}, \lv{y}) \vee \pred{F}(\lv{y}, \lv{x})
\end{align}

Unit propagation: 
\begin{align}
\forall \lv{y} \in \pop{q_T}, \lv{z} \in \pop{q_F}:&~ \neg \pred{F}(\lv{y}, \lv{z}) \\
\forall \lv{x} \in \pop{q_T}, \lv{z} \in \pop{q_T}:&~ \pred{F}(\lv{x}, \lv{z}) \\
\forall \lv{x} \in \pop{q_F}, \lv{y} \in \pop{q_T}:&~ \neg \pred{F}(\lv{x}, \lv{y}) \\
\forall \lv{x} \in \pop{q_F}, \lv{y} \in \pop{q_F}, \lv{z} \in \pop{q_F}:&~ \neg \pred{F}(\lv{x}, \lv{y}) \vee \neg \pred{F}(\lv{y}, \lv{z}) \vee \pred{F}(\lv{x}, \lv{z}) \\
\forall \lv{y} \in \pop{q_T}:&~ \pred{F}(\lv{y}, \const{N}) \\
\forall \lv{x} \in \pop{q_F}:&~ \neg \pred{F}(\lv{x}, \const{N}) \\
\forall \lv{x} \in \pop{q_F}, \lv{y} \in \pop{q_F}:&~ \neg \pred{F}(\lv{x}, \lv{y}) \vee \pred{F}(\lv{y}, \lv{x})
\end{align}

The first, second, third, fifth, and sixth clauses are independent of the other clauses and can be reasoned about separately. They can be trivially lifted. The remaining clauses are: 
\begin{align}
\forall \lv{x} \in \pop{q_F}, \lv{y} \in \pop{q_F}, \lv{z} \in \pop{q_F}:&~ \neg \pred{F}(\lv{x}, \lv{y}) \vee \neg \pred{F}(\lv{y}, \lv{z}) \vee \pred{F}(\lv{x}, \lv{z}) \\ 
\forall \lv{x} \in \pop{q_F}, \lv{y} \in \pop{q_F}:&~ \neg \pred{F}(\lv{x}, \lv{y}) \vee \pred{F}(\lv{y}, \lv{x})
\end{align}
The above clauses are an instance of our initial clauses but with smaller domain sizes. We can reason about them by following a similar process, and if we keep sub-results in a cache, the process will be polynomial.
\end{proof}

\subsection{Proof of Proposition~\ref{birth-para-prop}}
\begin{proof}
After skolemization \cite{van2014skolemization} for removing the existential quantifier, the birthday-paradox theory contains  
$\forall \lv{p} \in \pop{p}, \forall \lv{d} \in \pop{d}: \pred{S}(\lv{p}) \vee \neg \pred{Born}(\lv{p}, \lv{d})$,
$\forall \lv{p} \in \pop{p}, \lv{d}_1, \lv{d}_2 \in \pop{d}: \neg \pred{Born}(\lv{p}, \lv{d}_1) \vee \neg \pred{Born}(\lv{p}, \lv{d}_2)$, and
$\forall \lv{p}_1, \lv{p}_2 \in \pop{p}, \lv{d} \in \pop{d}: \neg \pred{Born}(\lv{p}_1, \lv{d}) \vee \neg \pred{Born}(\lv{p}_2, \lv{d})$, where $\pred{S}$ is the Skolem predicate. 
This theory is not in $\FOtwo$ and not in $RU$ and is not domain-liftable using \rules. However, this theory is both $\StwoFOtwo$ and $\StwoRU$ as the last two clauses are instances of $\alpha{Born}$, the first one is in $\FOtwo$ and also in $RU$ and has no PRVs with more than one LV other than $\pred{Born}$. Therefore, this theory is domain-liftable based on Theorem~\ref{liftable-class}.
\end{proof}

\subsection{Proof of Theorem~\ref{liftable-class}}
\begin{proof}
The case where $\alpha=\emptyset$ is trivial. Let $\alpha = \alpha(\pred{S}_1) \wedge \alpha(\pred{S}_2) \wedge \dots \wedge \alpha(\pred{S}_n)$. Once we remove all PRVs having none or one LV by (lifted) case-analysis, the remaining clauses can be divided into $n+1$ components: the $i$-th component in the first $n$ components only contains $\pred{S}_i$ literals, and the $(n+1)$-th component contains no $\pred{S}_i$ literals. These components are disconnected from each other, so we can consider each of them separately. The $(n+1)$-th component comes from clauses in $\beta$ and is domain-liftable by definition. 
The following two Lemmas prove that the clauses in the other components are also domain-liftable. The proofs of both lemmas rely on domain recursion.
\begin{lemma} \label{nine-clauses}
A clausal theory with only one predicate $\pred{S}$ is domain-liftable if all clauses have exactly two different literals of $\pred{S}$.
\end{lemma}

\begin{lemma} \label{with-unaries}
Suppose $\{\pop{p_1}, \pop{p_2}, \dots, \pop{p_n}\}$ are mutually exclusive subsets of $\pop{x}$ and $\{\pop{q_1}, \pop{q_2}, \dots, \pop{q_m}\}$ are mutually exclusive subsets of $\pop{y}$. We can add any unit clause of the form $\forall \lv{p}_i \in \pop{p_i}, \lv{q}_j \in \pop{q_j}: \pred{S}(\lv{p}_i, \lv{q}_j)$ or $\forall \lv{p}_i \in \pop{p_i}, \lv{q}_j \in \pop{q_j}: \neg \pred{S}(\lv{p}_i, \lv{q}_j)$ to the theory in Lemma~\ref{nine-clauses} and the theory is still domain-liftable.
\end{lemma}
Therefore, theories in $\StwoFOtwo$ and $\StwoRU$ are domain-liftable.
\end{proof}

\subsection{Proof of Lemma~\ref{nine-clauses}}
\begin{proof}
A theory in this form has a \emph{subset} of the following clauses:
\begin{align}
\forall \lv{x} \in \pop{x}, \lv{y_1}, \lv{y_2} \in \pop{y}:&~ \pred{S}(\lv{x}, \lv{y_1}) \vee \pred{S}(\lv{x}, \lv{y_2}) \\
\forall \lv{x} \in \pop{x}, \lv{y_1}, \lv{y_2} \in \pop{y}:&~ \pred{S}(\lv{x}, \lv{y_1}) \vee \neg \pred{S}(\lv{x}, \lv{y_2}) \\
\forall \lv{x} \in \pop{x}, \lv{y_1}, \lv{y_2} \in \pop{y}:&~ \neg \pred{S}(\lv{x}, \lv{y_1}) \vee \neg \pred{S}(\lv{x}, \lv{y_2}) \\
\forall \lv{x_1}, \lv{x_2} \in \pop{x}, \lv{y} \in \pop{y}:&~ \pred{S}(\lv{x_1},\lv{y}) \vee \pred{S}(\lv{x_2},\lv{y}) \\
\forall \lv{x_1}, \lv{x_2} \in \pop{x}, \lv{y} \in \pop{y}:&~ \pred{S}(\lv{x_1},\lv{y}) \vee \neg \pred{S}(\lv{x_2},\lv{y}) \\
\forall \lv{x_1}, \lv{x_2} \in \pop{x}, \lv{y} \in \pop{y}:&~ \neg \pred{S}(\lv{x_1},\lv{y}) \vee \neg \pred{S}(\lv{x_2},\lv{y}) \\
\forall \lv{x_1}, \lv{x_2} \in \pop{x}, \lv{y_1}, \lv{y_2} \in \pop{y}:&~ \pred{S}(\lv{x_1}, \lv{y_1}) \vee \pred{S}(\lv{x_2}, \lv{y_2}) \\
\forall \lv{x_1}, \lv{x_2} \in \pop{x}, \lv{y_1}, \lv{y_2} \in \pop{y}:&~ \pred{S}(\lv{x_1}, \lv{y_1}) \vee \neg \pred{S}(\lv{x_2}, \lv{y_2}) \\
\forall \lv{x_1}, \lv{x_2} \in \pop{x}, \lv{y_1}, \lv{y_2} \in \pop{y}:&~ \neg \pred{S}(\lv{x_1}, \lv{y_1}) \vee \neg \pred{S}(\lv{x_2}, \lv{y_2}) 
\end{align}

Let $\const{N}$ be an individual in $\pop{x}$. Applying domain recursion on $\pop{x'}=\pop{x'} - \{\const{N}\} $ for all clauses gives:\\

for (1):
\begin{align}
\forall \lv{y_1}, \lv{y_2} \in \pop{y}:&~ \pred{S}(\const{N}, \lv{y_1}) \vee \pred{S}(\const{N}, \lv{y_2})\\
\forall \lv{x} \in \pop{x'}, \lv{y_1}, \lv{y_2} \in \pop{y}:&~ \pred{S}(\lv{x}, \lv{y_1}) \vee \pred{S}(\lv{x}, \lv{y_2})
\end{align}

for (2):
\begin{align}
\forall \lv{y_1}, \lv{y_2} \in \pop{y}: \pred{S}(\const{N}, \lv{y_1}) \vee \neg \pred{S}(\const{N}, \lv{y_2})\\
\forall \lv{x} \in \pop{x'}, \lv{y_1}, \lv{y_2} \in \pop{y}: \pred{S}(\lv{x}, \lv{y_1}) \vee \neg \pred{S}(\lv{x}, \lv{y_2})
\end{align}

for (3):
\begin{align}
\forall \lv{y_1}, \lv{y_2} \in \pop{y}: \neg \pred{S}(\const{N}, \lv{y_1}) \vee \neg \pred{S}(\const{N}, \lv{y_2})\\
\forall \lv{x} \in \pop{x'}, \lv{y_1}, \lv{y_2} \in \pop{y}: \neg \pred{S}(\lv{x}, \lv{y_1}) \vee \neg \pred{S}(\lv{x}, \lv{y_2})
\end{align}

for (4):
\begin{align}
\forall \lv{x} \in \pop{x'}, \lv{y} \in \pop{y}:&~ \pred{S}(\const{N},\lv{y}) \vee \pred{S}(\lv{x},\lv{y})\\
\forall \lv{x_1}, \lv{x_2} \in \pop{x'}, \lv{y} \in \pop{y}:&~ \pred{S}(\lv{x_1},\lv{y}) \vee \pred{S}(\lv{x_2},\lv{y})
\end{align}

for (5):
\begin{align}
\forall \lv{x} \in \pop{x'}, \lv{y} \in \pop{y}:&~ \pred{S}(\const{N},\lv{y}) \vee \neg \pred{S}(\lv{x},\lv{y})\\
\forall \lv{x} \in \pop{x'}, \lv{y} \in \pop{y}:&~ \neg \pred{S}(\const{N},\lv{y}) \vee \pred{S}(\lv{x},\lv{y})\\
\forall \lv{x_1}, \lv{x_2} \in \pop{x'}, \lv{y} \in \pop{y}:&~ \pred{S}(\lv{x_1},\lv{y}) \vee \neg \pred{S}(\lv{x_2},\lv{y})
\end{align}

for (6): 
\begin{align}
\forall \lv{x} \in \pop{x'}, \lv{y} \in \pop{y}:&~ \neg \pred{S}(\const{N},\lv{y}) \vee \neg \pred{S}(\lv{x},\lv{y})\\
\forall \lv{x_1}, \lv{x_2} \in \pop{x'}, \lv{y} \in \pop{y}:&~ \neg \pred{S}(\lv{x_1},\lv{y}) \vee \neg \pred{S}(\lv{x_2},\lv{y})
\end{align}

for (7):
\begin{align}
\forall \lv{x} \in \pop{x'}, \lv{y_1}, \lv{y_2} \in \pop{y}:&~ \pred{S}(\const{N}, \lv{y_1}) \vee \pred{S}(\lv{x}, \lv{y_2})\\
\forall \lv{x_1},\lv{x_2} \in \pop{x'}, \lv{y_1}, \lv{y_2} \in \pop{y}:&~ \pred{S}(\lv{x_1}, \lv{y_1}) \vee \pred{S}(\lv{x_2}, \lv{y_2})
\end{align}

for (8):
\begin{align}
\forall \lv{x} \in \pop{x'}, \lv{y_1}, \lv{y_2} \in \pop{y}:&~ \pred{S}(\const{N}, \lv{y_1}) \vee \neg \pred{S}(\lv{x}, \lv{y_2}) \\
\forall \lv{x} \in \pop{x'}, \lv{y_1}, \lv{y_2} \in \pop{y}:&~ \neg \pred{S}(\const{N}, \lv{y_1}) \vee \pred{S}(\lv{x}, \lv{y_2}) \\
\forall \lv{x_1}, \lv{x_2} \in \pop{x'}, \lv{y_1}, \lv{y_2} \in \pop{y}:&~ \pred{S}(\lv{x_1}, \lv{y_1}) \vee \neg \pred{S}(\lv{x_2}, \lv{y_2})
\end{align}

for (9):
\begin{align}
\forall \lv{x} \in \pop{x'}, \lv{y_1}, \lv{y_2} \in \pop{y}: \neg \pred{S}(\const{N}, \lv{y_1}) \vee \neg \pred{S}(\lv{x}, \lv{y_2})\\
\forall \lv{x_1},\lv{x_2} \in \pop{x'}, \lv{y_1}, \lv{y_2} \in \pop{y}: \neg \pred{S}(\lv{x_1}, \lv{y_1}) \vee \neg \pred{S}(\lv{x_2}, \lv{y_2})
\end{align}

Then we can perform lifted case-analysis on $\pred{S}(\const{N},\lv{y})$. For the case where $\pred{S}(\const{N},\lv{y})$ is true for exactly $k$ of the individuals in $\pop{y}$, we update all clauses assuming $\pop{y_T}$ and $\pop{y_F}$ represent the individuals for which $\pred{S}(\const{N},\lv{y})$ is \true\ and \false\ respectively, and assuming $\forall \lv{y} \in \pop{y_T}: \pred{S}(\const{N},\lv{y})$ and $\forall \lv{y} \in \pop{y_F}: \neg \pred{S}(\const{N},\lv{y})$:\\

for (1):
\begin{align}
\forall \lv{x} \in \pop{x'}, \lv{y_1}, \lv{y_2} \in \pop{y}: \pred{S}(\lv{x}, \lv{y_1}) \vee \pred{S}(\lv{x}, \lv{y_2})
\end{align}

for (2):
\begin{align}
\forall \lv{x} \in \pop{x'}, \lv{y_1}, \lv{y_2} \in \pop{y}: \pred{S}(\lv{x}, \lv{y_1}) \vee \neg \pred{S}(\lv{x}, \lv{y_2})
\end{align}

for (3):
\begin{align}
\forall \lv{x} \in \pop{x'}, \lv{y_1}, \lv{y_2} \in \pop{y}: \neg \pred{S}(\lv{x}, \lv{y_1}) \vee \neg \pred{S}(\lv{x}, \lv{y_2})
\end{align}

for (4): 
\begin{align}
\forall \lv{x} \in \pop{x'}, \lv{y} \in \pop{y_F}:&~ \pred{S}(\lv{x},\lv{y})\\
\forall \lv{x_1}, \lv{x_2} \in \pop{x'}, \lv{y} \in \pop{y}:&~ \pred{S}(\lv{x_1},\lv{y}) \vee \pred{S}(\lv{x_2},\lv{y})
\end{align}

for (5):
\begin{align}
\forall \lv{x} \in \pop{x'}, \lv{y} \in \pop{y_F}:&~ \neg \pred{S}(\lv{x},\lv{y})\\
\forall \lv{x} \in \pop{x'}, \lv{y} \in \pop{y_T}:&~ \pred{S}(\lv{x},\lv{y})\\
\forall \lv{x_1}, \lv{x_2} \in \pop{x'}, \lv{y} \in \pop{y}:&~ \pred{S}(\lv{x_1},\lv{y}) \vee \neg \pred{S}(\lv{x_2},\lv{y})
\end{align}

for (6): 
\begin{align}
\forall \lv{x} \in \pop{x'}, \lv{y} \in \pop{y_T}:&~ \neg \pred{S}(\lv{x},\lv{y})\\
\forall \lv{x_1}, \lv{x_2} \in \pop{x'}, \lv{y} \in \pop{y}:&~ \neg \pred{S}(\lv{x_1},\lv{y}) \vee \neg \pred{S}(\lv{x_2},\lv{y})
\end{align}

for (7):
\begin{align}
\forall \lv{x} \in \pop{x'}, \lv{y_1} \in \pop{y_F}, \lv{y_2} \in \pop{y}:&~ \pred{S}(\lv{x}, \lv{y_2})\\
\forall \lv{x_1}, \lv{x_2} \in \pop{x'}, \lv{y_1}, \lv{y_2} \in \pop{y}:&~ \pred{S}(\lv{x_1}, \lv{y_1}) \vee \pred{S}(\lv{x_2}, \lv{y_2})
\end{align}

for (8):
\begin{align}
\forall \lv{x} \in \pop{x'}, \lv{y_1} \in \pop{y_F}, \lv{y_2} \in \pop{y}:&~ \neg \pred{S}(\lv{x}, \lv{y_2}) \\
\forall \lv{x} \in \pop{x'}, \lv{y_1} \in \pop{y_T}, \lv{y_2} \in \pop{y}:&~ \pred{S}(\lv{x}, \lv{y_2}) \\
\forall \lv{x_1}, \lv{x_2} \in \pop{x'}, \lv{y_1}, \lv{y_2} \in \pop{y}:&~ \pred{S}(\lv{x_1}, \lv{y_1}) \vee \neg \pred{S}(\lv{x_2}, \lv{y_2})\\
\end{align}

for (9):
\begin{align}
\forall \lv{x} \in \pop{x'}, \lv{y_1} \in \pop{y_T}, \lv{y_2} \in \pop{y}:&~ \neg \pred{S}(\lv{x}, \lv{y_2})\\
\forall \lv{x_1}, \lv{x_2} \in \pop{x'}, \lv{y_1}, \lv{y_2} \in \pop{y}:&~ \neg \pred{S}(\lv{x_1}, \lv{y_1}) \vee \neg \pred{S}(\lv{x_2}, \lv{y_2})
\end{align}

After subsumptions and shattering: \\

for (1):
\begin{align}
\forall \lv{x} \in \pop{x'}, \lv{y_1}, \lv{y_2} \in \pop{y_T}:&~ \pred{S}(\lv{x}, \lv{y_1}) \vee \pred{S}(\lv{x}, \lv{y_2})\\
\forall \lv{x} \in \pop{x'}, \lv{y_1} \in \pop{y_T}, \lv{y_2} \in \pop{y_F}:&~ \pred{S}(\lv{x}, \lv{y_1}) \vee \pred{S}(\lv{x}, \lv{y_2})\\
\forall \lv{x} \in \pop{x'}, \lv{y_1} \in \pop{y_F}, \lv{y_2} \in \pop{y_T}:&~ \pred{S}(\lv{x}, \lv{y_1}) \vee \pred{S}(\lv{x}, \lv{y_2})\\
\forall \lv{x} \in \pop{x'}, \lv{y_1}, \lv{y_2} \in \pop{y_F}:&~ \pred{S}(\lv{x}, \lv{y_1}) \vee \pred{S}(\lv{x}, \lv{y_2})\\
\end{align}

for (2):
\begin{align}
\forall \lv{x} \in \pop{x'}, \lv{y_1}, \lv{y_2} \in \pop{y_T}:&~ \pred{S}(\lv{x}, \lv{y_1}) \vee \neg \pred{S}(\lv{x}, \lv{y_2})\\
\forall \lv{x} \in \pop{x'}, \lv{y_1} \in \pop{y_T}, \lv{y_2} \in \pop{y_F}:&~ \pred{S}(\lv{x}, \lv{y_1}) \vee \neg \pred{S}(\lv{x}, \lv{y_2})\\
\forall \lv{x} \in \pop{x'}, \lv{y_1} \in \pop{y_F}, \lv{y_2} \in \pop{y_T}:&~ \pred{S}(\lv{x}, \lv{y_1}) \vee \neg \pred{S}(\lv{x}, \lv{y_2})\\
\forall \lv{x} \in \pop{x'}, \lv{y_1} \in \pop{y_F}, \lv{y_2} \in \pop{y_F}:&~ \pred{S}(\lv{x}, \lv{y_1}) \vee \neg \pred{S}(\lv{x}, \lv{y_2})
\end{align}

for (3): 
\begin{align}
\forall \lv{x} \in \pop{x'}, \lv{y_1}, \lv{y_2} \in \pop{y_T}:&~ \neg \pred{S}(\lv{x}, \lv{y_1}) \vee \neg \pred{S}(\lv{x}, \lv{y_2})\\
\forall \lv{x} \in \pop{x'}, \lv{y_1} \in \pop{y_T}, \lv{y_2} \in \pop{y_F}:&~ \neg \pred{S}(\lv{x}, \lv{y_1}) \vee \neg \pred{S}(\lv{x}, \lv{y_2})\\
\forall \lv{x} \in \pop{x'}, \lv{y_1} \in \pop{y_F}, \lv{y_2} \in \pop{y_T}:&~ \neg \pred{S}(\lv{x}, \lv{y_1}) \vee \neg \pred{S}(\lv{x}, \lv{y_2})\\
\forall \lv{x} \in \pop{x'}, \lv{y_1} \in \pop{y_F}, \lv{y_2} \in \pop{y_F}:&~ \neg \pred{S}(\lv{x}, \lv{y_1}) \vee \neg \pred{S}(\lv{x}, \lv{y_2})
\end{align}

for (4): (For the second clause, the case where $\lv{y} \in \pop{y_F}$ becomes subsumed by the first clause)
\begin{align}
\forall \lv{x} \in \pop{x'}, \lv{y} \in \pop{y_F}:&~ \pred{S}(\lv{x},\lv{y})\\
\forall \lv{x_1}, \lv{x_2} \in \pop{x'}, \lv{y} \in \pop{y_T}:&~ \pred{S}(\lv{x_1},\lv{y}) \vee \pred{S}(\lv{x_2},\lv{y})
\end{align}

for (5): (The third clause was subsumed by the first two)
\begin{align}
\forall \lv{x} \in \pop{x'}, \lv{y} \in \pop{y_F}:&~ \neg \pred{S}(\lv{x},\lv{y})\\
\forall \lv{x} \in \pop{x'}, \lv{y} \in \pop{y_T}:&~ \pred{S}(\lv{x},\lv{y})
\end{align}

for (6): (Similar to (4))
\begin{align}
\forall \lv{x} \in \pop{x'}, \lv{y} \in \pop{y_T}:&~ \neg \pred{S}(\lv{x},\lv{y})\\
\forall \lv{x_1}, \lv{x_2} \in \pop{x'}, \lv{y} \in \pop{y_F}:&~ \neg \pred{S}(\lv{x_1},\lv{y}) \vee \neg \pred{S}(\lv{x_2},\lv{y})
\end{align}

for (7): 
\begin{align}
\forall \lv{x} \in \pop{x'}, \lv{y_1} \in \pop{y_F}, \lv{y_2} \in \pop{y_T}:&~ \pred{S}(\lv{x}, \lv{y_2})\\
\forall \lv{x} \in \pop{x'}, \lv{y_1} \in \pop{y_F}, \lv{y_2} \in \pop{y_F}:&~ \pred{S}(\lv{x}, \lv{y_2})\\
\forall \lv{x_1},\lv{x_2} \in \pop{x'}, \lv{y_1} \in \pop{y_T}, \lv{y_2} \in \pop{y_T}:&~ \pred{S}(\lv{x_1}, \lv{y_1}) \vee \pred{S}(\lv{x_2}, \lv{y_2})\\
\forall \lv{x_1},\lv{x_2} \in \pop{x'}, \lv{y_1} \in \pop{y_T}, \lv{y_2} \in \pop{y_F}:&~ \pred{S}(\lv{x_1}, \lv{y_1}) \vee \pred{S}(\lv{x_2}, \lv{y_2})
\end{align}

for (8): 
\begin{align}
\forall \lv{x} \in \pop{x'}, \lv{y_1} \in \pop{y_F}, \lv{y_2} \in \pop{y_T}:&~ \neg \pred{S}(\lv{x}, \lv{y_2}) \\
\forall \lv{x} \in \pop{x'}, \lv{y_1} \in \pop{y_F}, \lv{y_2} \in \pop{y_F}:&~ \neg \pred{S}(\lv{x}, \lv{y_2}) \\
\forall \lv{x} \in \pop{x'}, \lv{y_1} \in \pop{y_T}, \lv{y_2} \in \pop{y_T}:&~ \pred{S}(\lv{x}, \lv{y_2}) \\
\forall \lv{x} \in \pop{x'}, \lv{y_1} \in \pop{y_T}, \lv{y_2} \in \pop{y_F}:&~ \pred{S}(\lv{x}, \lv{y_2})
\end{align}

for (9): 
\begin{align}
\forall \lv{x} \in \pop{x'}, \lv{y_1} \in \pop{y_T}, \lv{y_2} \in \pop{y_T}:&~ \neg \pred{S}(\lv{x}, \lv{y_2})\\
\forall \lv{x} \in \pop{x'}, \lv{y_1} \in \pop{y_T}, \lv{y_2} \in \pop{y_F}:&~ \neg \pred{S}(\lv{x}, \lv{y_2})\\
\forall \lv{x_1},\lv{x_2} \in \pop{x'}, \lv{y_1} \in \pop{y_F}, \lv{y_2} \in \pop{y_T}:&~ \neg \pred{S}(\lv{x_1}, \lv{y_1}) \vee \neg \pred{S}(\lv{x_2}, \lv{y_2})\\
\forall \lv{x_1},\lv{x_2} \in \pop{x'}, \lv{y_1} \in \pop{y_F}, \lv{y_2} \in \pop{y_F}:&~ \neg \pred{S}(\lv{x_1}, \lv{y_1}) \vee \neg \pred{S}(\lv{x_2}, \lv{y_2})
\end{align}

Looking at the first clause for (7) (and some other clauses), we see that there exists a $\forall \lv{y_1} \in \pop{y_T}$ but $\lv{y_1}$ does not appear in the formula. If $\pop{y_T}=\emptyset$, we can ignore this clause. Otherwise, we can ignore $\forall \lv{y_1} \in \pop{y_T}$. So we consider three cases. When $k=0$ (i.e. $\pop{y_F} = \pop{y}$, $\pop{y_T} = \emptyset$): \\

for (1):
\begin{align}
\forall \lv{x} \in \pop{x'}, \lv{y_1}, \lv{y_2} \in \pop{y}:&~ \pred{S}(\lv{x}, \lv{y_1}) \vee \pred{S}(\lv{x}, \lv{y_2})
\end{align}

for (2):
\begin{align}
\forall \lv{x} \in \pop{x'}, \lv{y_1}, \lv{y_2} \in \pop{y}:&~ \pred{S}(\lv{x}, \lv{y_1}) \vee \neg \pred{S}(\lv{x}, \lv{y_2})
\end{align}

for (3):
\begin{align}
\forall \lv{x} \in \pop{x'}, \lv{y_1}, \lv{y_2} \in \pop{y}:&~ \neg \pred{S}(\lv{x}, \lv{y_1}) \vee \neg \pred{S}(\lv{x}, \lv{y_2})
\end{align}

for (4):
\begin{align}
\forall \lv{x} \in \pop{x'}, \lv{y} \in \pop{y}:&~ \pred{S}(\lv{x},\lv{y})
\end{align}

for (5):
\begin{align}
\forall \lv{x} \in \pop{x'}, \lv{y} \in \pop{y}:&~ \neg \pred{S}(\lv{x},\lv{y})
\end{align}

for (6):
\begin{align}
\forall \lv{x_1}, \lv{x_2} \in \pop{x'}, \lv{y} \in \pop{y}:&~ \neg \pred{S}(\lv{x_1},\lv{y}) \vee \neg \pred{S}(\lv{x_2},\lv{y})
\end{align}

for (7):
\begin{align}
\forall \lv{x} \in \pop{x'}, \lv{y_2} \in \pop{y}:&~ \pred{S}(\lv{x}, \lv{y_2})
\end{align}

for (8):
\begin{align}
\forall \lv{x} \in \pop{x'}, \lv{y_2} \in \pop{y}:&~ \neg \pred{S}(\lv{x}, \lv{y_2})
\end{align}

for (9):\begin{align}
\forall \lv{x_1},\lv{x_2} \in \pop{x'}, \lv{y_1} \in \pop{y}, \lv{y_2} \in \pop{y}:&~ \neg \pred{S}(\lv{x_1}, \lv{y_1}) \vee \neg \pred{S}(\lv{x_2}, \lv{y_2})
\end{align}

If clause \#4 is one of the clauses in the theory, then unit propagation either gives \false, or satisfies all the clauses. The same is true for clauses \#5, \#7, and \#8. In a theory not having any of these four clauses, we will be left with a set of clauses that are again a subset of the initial 9 clauses that we started with, but with a smaller domain size. By applying the same procedure, we can count the number of models.
When $k=|\pop{y}|$ (i.e. $\pop{y_F} = \emptyset$, $\pop{y_T} = \pop{y}$), everything is just similar to the $k=0$ case. \\

When $0<k<|\pop{y}|$ (i.e. neither $\pop{y_T}$ nor $\pop{y_F}$ are empty): \\

for (1):
\begin{align}
\forall \lv{x} \in \pop{x'}, \lv{y_1}, \lv{y_2} \in \pop{y_T}:&~ \pred{S}(\lv{x}, \lv{y_1}) \vee \pred{S}(\lv{x}, \lv{y_2})\\
\forall \lv{x} \in \pop{x'}, \lv{y_1} \in \pop{y_T}, \lv{y_2} \in \pop{y_F}:&~ \pred{S}(\lv{x}, \lv{y_1}) \vee \pred{S}(\lv{x}, \lv{y_2})\\
\forall \lv{x} \in \pop{x'}, \lv{y_1} \in \pop{y_F}, \lv{y_2} \in \pop{y_T}:&~ \pred{S}(\lv{x}, \lv{y_1}) \vee \pred{S}(\lv{x}, \lv{y_2})\\
\forall \lv{x} \in \pop{x'}, \lv{y_1}, \lv{y_2} \in \pop{y_F}:&~ \pred{S}(\lv{x}, \lv{y_1}) \vee \pred{S}(\lv{x}, \lv{y_2})
\end{align}

for (2):
\begin{align}
\forall \lv{x} \in \pop{x'}, \lv{y_1}, \lv{y_2} \in \pop{y_T}:&~ \pred{S}(\lv{x}, \lv{y_1}) \vee \neg \pred{S}(\lv{x}, \lv{y_2})\\
\forall \lv{x} \in \pop{x'}, \lv{y_1} \in \pop{y_T}, \lv{y_2} \in  \pop{y_F}:&~ \pred{S}(\lv{x}, \lv{y_1}) \vee \neg \pred{S}(\lv{x}, \lv{y_2})\\
\forall \lv{x} \in \pop{x'}, \lv{y_1} \in \pop{y_F}, \lv{y_2} \in \pop{y_T}:&~ \pred{S}(\lv{x}, \lv{y_1}) \vee \neg \pred{S}(\lv{x}, \lv{y_2})\\
\forall \lv{x} \in \pop{x'}, \lv{y_1} \in \pop{y_F}, \lv{y_2} \in \pop{y_F}:&~ \pred{S}(\lv{x}, \lv{y_1}) \vee \neg \pred{S}(\lv{x}, \lv{y_2})
\end{align}

for (3):
\begin{align}
\forall \lv{x} \in \pop{x'}, \lv{y_1}, \lv{y_2} \in \pop{y_T}:&~ \neg \pred{S}(\lv{x}, \lv{y_1}) \vee \neg \pred{S}(\lv{x}, \lv{y_2})\\
\forall \lv{x} \in \pop{x'}, \lv{y_1} \pop{y_T}, \lv{y_2} \in \pop{y_F}:&~ \neg \pred{S}(\lv{x}, \lv{y_1}) \vee \neg \pred{S}(\lv{x}, \lv{y_2})\\
\forall \lv{x} \in \pop{x'}, \lv{y_1} \in \pop{y_F}, \lv{y_2} \in \pop{y_T}:&~ \neg \pred{S}(\lv{x}, \lv{y_1}) \vee \neg \pred{S}(\lv{x}, \lv{y_2})\\
\forall \lv{x} \in \pop{x'}, \lv{y_1} \in \pop{y_F}, \lv{y_2} \in \pop{y_F}:&~ \neg \pred{S}(\lv{x}, \lv{y_1}) \vee \neg \pred{S}(\lv{x}, \lv{y_2})
\end{align}

for (4):
\begin{align}
\forall \lv{x} \in \pop{x'}, \lv{y} \in \pop{y_F}:&~ \pred{S}(\lv{x},\lv{y})\\
\forall \lv{x_1}, \lv{x_2} \in \pop{x'}, \lv{y} \in \pop{y_T}:&~ \pred{S}(\lv{x_1},\lv{y}) \vee \pred{S}(\lv{x_2},\lv{y})
\end{align}

for (5):
\begin{align}
\forall \lv{x} \in \pop{x'}, \lv{y} \in \pop{y_F}:&~ \neg \pred{S}(\lv{x},\lv{y})\\
\forall \lv{x} \in \pop{x'}, \lv{y} \in \pop{y_T}:&~ \pred{S}(\lv{x},\lv{y})
\end{align}

for (6):
\begin{align}
\forall \lv{x} \in \pop{x'}, \lv{y} \in \pop{y_T}:&~ \neg \pred{S}(\lv{x},\lv{y})\\
\forall \lv{x_1}, \lv{x_2} \in \pop{x'}, \lv{y} \in \pop{y_F}:&~ \neg \pred{S}(\lv{x_1},\lv{y}) \vee \neg \pred{S}(\lv{x_2},\lv{y})
\end{align}

for (7):
\begin{align}
\forall \lv{x} \in \pop{x'}, \lv{y} \in \pop{y_T}:&~ \pred{S}(\lv{x},\lv{y})\\
\forall \lv{x} \in \pop{x'}, \lv{y} \in \pop{y_F}:&~ \pred{S}(\lv{x},\lv{y})
\end{align}

for (8):
\begin{align}
False
\end{align}

for (9):
\begin{align}
\forall \lv{x} \in \pop{x'}, \lv{y} \in \pop{y_T}:&~ \neg \pred{S}(\lv{x},\lv{y})\\
\forall \lv{x} \in \pop{x'}, \lv{y} \in \pop{y_F}:&~ \neg \pred{S}(\lv{x},\lv{y})
\end{align}

If either one of clauses \#5, \#7, \#8 or \#9 are in the theory, then unit propagation either gives $False$ or satisfies all clauses. Assume none of these four clauses are in the theory. If both clauses \#4 and \#6 are in a theory, again unit propagation gives either $False$ or satisfies all clauses. If none of them are in the theory, then the other clauses are a subset of the initial 9 clauses that we started with. So let's consider the case where we have clause \#4 and a subset of the first three clauses (the case with \#6 instead of \#4 is similar). In this case, if clauses \#2 or \#3 are in the theory, unit propagation either gives $False$ or satisfies all the clauses. If none of them are in the theory and only \#1 is in the theory, we will have the following clauses after unit propagation:
\begin{align}
\forall \lv{x} \in \pop{x'}, \lv{y} \in \pop{y_F}:&~ \pred{S}(\lv{x},\lv{y})\\
\forall \lv{x_1}, \lv{x_2} \in \pop{x'}, \lv{y} \in \pop{y_T}:&~ \pred{S}(\lv{x_1},\lv{y}) \vee \pred{S}(\lv{x_2},\lv{y})\\
\forall \lv{x} \in \pop{x'}, \lv{y_1}, \lv{y_2} \in \pop{y_T}:&~ \pred{S}(\lv{x}, \lv{y_1}) \vee \pred{S}(\lv{x}, \lv{y_2})
\end{align}
The first clause is independent of the other two clauses. The second and third clauses are just similar to clauses \#4 and \#1 in the initial list of clauses and we can handle them using the same procedure.

If we use a cache to store computations for all subproblems, WFOMC is domain-liftable, i.e. polynomial in the population sizes.
\end{proof}

\subsection{Proof of Lemma~\ref{with-unaries}}
\begin{proof}
Let $\psi$ be the set of pairs $(i,j)$ such that the singleton clause $\forall \lv{p_i} \in \pop{p_i}, \lv{q_j} \in \pop{q_j}: \pred{S}(\lv{p_i}, \lv{q_j})$ is in the theory, and $\overline{\psi}$ be the set of pairs $(i,j)$ such that the singleton clause $\forall \lv{p_i} \in \pop{p_i}, \lv{q_j} \in \pop{q_j}: \neg \pred{S}(\lv{p_i}, \lv{q_j})$ is in the theory. Then the singleton clauses can be written as follows:
\begin{align}
\forall (i, j) \in \psi: \forall \lv{p_i} \in \pop{p_i}, \lv{q_j} \in \pop{q_j}:&~ \pred{S}(\lv{p_i}, \lv{q_j}) \\
\forall (i, j) \in \overline{\psi}: \forall \lv{p_i} \in \pop{p_i}, \lv{q_j} \in \pop{q_j}:&~ \neg \pred{S}(\lv{p_i}, \lv{q_j})
\end{align}
And the 2S clauses are as in Lemma~\ref{nine-clauses}. Without loss of generality, let's assume we select an individual $N \in \pop{p_1}$ for domain recursion, and re-write all clauses to separate $N$ from $\pop{p_1}$. Assuming $\pop{p'_1},=\pop{p_1}-\{\const{N}\}$ and $\pop{x'}=\pop{x}-\{\const{N}\}$, the theory will be: \\

For singletons: 
\begin{align}
\forall (1, j) \in \psi: \forall \lv{q_j} \in \pop{q_j}:&~ \pred{S}(\const{N}, \lv{q_j}) \\
\forall (1, j) \in \overline{\psi}: \forall \lv{q_j} \in \pop{q_j}:&~ \neg \pred{S}(\const{N}, \lv{q_j}) \\
\forall (1, j) \in \psi: \forall \lv{p'_1} \in \pop{p'_1},, \lv{q_j} \in \pop{q_j}:&~ \pred{S}(\lv{p'_1}, \lv{q_j}) \\
\forall (1, j) \in \overline{\psi}: \forall \lv{p'_1} \in \pop{p'_1},, \lv{q_j} \in \pop{q_j}:&~ \neg \pred{S}(\lv{p'_1}, \lv{q_j}) \\
\forall (i \neq 1, j) \in \psi: \forall \lv{p_i} \in \pop{p_i}, \lv{q_j} \in \pop{q_j}:&~ \pred{S}(\lv{p_i}, \lv{q_j}) \\
\forall (i \neq 1, j) \in \overline{\psi}: \forall \lv{p_i} \in \pop{p_i}, \lv{q_j} \in \pop{q_j}:&~ \neg \pred{S}(\lv{p_i}, \lv{q_j}) 
\end{align}

for (1):
\begin{align}
\forall \lv{y_1}, \lv{y_2} \in \pop{y}:&~ \pred{S}(\const{N}, \lv{y_1}) \vee \pred{S}(\const{N}, \lv{y_2})\\
\forall \lv{x} \in \pop{x'}, \lv{y_1}, \lv{y_2} \in \pop{y}:&~ \pred{S}(\lv{x}, \lv{y_1}) \vee \pred{S}(\lv{x}, \lv{y_2})
\end{align}

for (2):
\begin{align}
\forall \lv{y_1}, \lv{y_2} \in \pop{y}:&~ \pred{S}(\const{N}, \lv{y_1}) \vee \neg \pred{S}(\const{N}, \lv{y_2})\\
\forall \lv{x} \in \pop{x'}, \lv{y_1}, \lv{y_2} \in \pop{y}:&~ \pred{S}(\lv{x}, \lv{y_1}) \vee \neg \pred{S}(\lv{x}, \lv{y_2})
\end{align}

for (3): 
\begin{align}
\forall \lv{y_1}, \lv{y_2} \in \pop{y}:&~ \neg \pred{S}(\const{N}, \lv{y_1}) \vee \neg \pred{S}(\const{N}, \lv{y_2})\\
\forall \lv{x} \in \pop{x'}, \lv{y_1}, \lv{y_2} \in \pop{y}:&~ \neg \pred{S}(\lv{x}, \lv{y_1}) \vee \neg \pred{S}(\lv{x}, \lv{y_2})
\end{align}

for (4):
\begin{align}
\forall \lv{x} \in \pop{x'}, \lv{y} \in \pop{y}: \pred{S}(\const{N}, \lv{y}) \vee \pred{S}(\lv{x}, \lv{y})\\
\forall \lv{x_1}, \lv{x_2} \in \pop{x'}, \lv{y} \in \pop{y}: \pred{S}(\lv{x_1}, \lv{y}) \vee \pred{S}(\lv{x_2}, \lv{y})
\end{align}

for (5):
\begin{align}
\forall \lv{x} \in \pop{x'}, \lv{y} \in \pop{y}:&~ \pred{S}(\const{N}, \lv{y}) \vee \neg \pred{S}(\lv{x}, \lv{y})\\
\forall \lv{x} \in \pop{x'}, \lv{y} \in \pop{y}:&~ \neg \pred{S}(\const{N}, \lv{y}) \vee \pred{S}(\lv{x}, \lv{y})\\
\forall \lv{x_1}, \lv{x_2} \in \pop{x'}, \lv{y} \in \pop{y}:&~ \pred{S}(\lv{x_1}, \lv{y}) \vee \neg \pred{S}(\lv{x_2}, \lv{y})
\end{align}

for (6):
\begin{align}
\forall \lv{x} \in \pop{x'}, \lv{y} \in \pop{y}:&~ \neg \pred{S}(\const{N}, \lv{y}) \vee \neg \pred{S}(\lv{x}, \lv{y})\\
\forall \lv{x_1}, \lv{x_2} \in \pop{x'}, \lv{y} \in \pop{y}:&~ \neg \pred{S}(\lv{x_1}, \lv{y}) \vee \neg \pred{S}(\lv{x_2}, \lv{y})
\end{align}

for (7):
\begin{align}
\forall \lv{x} \in \pop{x'}, \lv{y_1}, \lv{y_2} \in \pop{y}:&~ \pred{S}(\const{N}, \lv{y_1}) \vee \pred{S}(\lv{x}, \lv{y_2})\\
\forall \lv{x_1},\lv{x_2} \in \pop{x'}, \lv{y_1}, \lv{y_2} \in \pop{y}:&~ \pred{S}(\lv{x_1}, \lv{y_1}) \vee \pred{S}(\lv{x_2}, \lv{y_2})
\end{align}

for (8):
\begin{align}
\forall \lv{x} \in \pop{x'}, \lv{y_1}, \lv{y_2} \in \pop{y}:&~ \pred{S}(\const{N}, \lv{y_1}) \vee \neg \pred{S}(\lv{x}, \lv{y_2}) \\
\forall \lv{x} \in \pop{x'}, \lv{y_1}, \lv{y_2} \in \pop{y}:&~ \neg \pred{S}(\const{N}, \lv{y_1}) \vee \pred{S}(\lv{x}, \lv{y_2}) \\
\forall \lv{x_1}, \lv{x_2} \in \pop{x'}, \lv{y_1}, \lv{y_2} \in \pop{y}:&~ \pred{S}(\lv{x_1}, \lv{y_1}) \vee \neg \pred{S}(\lv{x_2}, \lv{y_2})
\end{align}

for (9):
\begin{align}
\forall \lv{x} \in \pop{x'}, \lv{y_1}, \lv{y_2} \in \pop{y}:&~ \neg \pred{S}(\const{N}, \lv{y_1}) \vee \neg \pred{S}(\lv{x}, \lv{y_2})\\
\forall \lv{x_1},\lv{x_2} \in \pop{x'}, \lv{y_1}, \lv{y_2} \in \pop{y}:&~ \neg \pred{S}(\lv{x_1}, \lv{y_1}) \vee \neg \pred{S}(\lv{x_2}, \lv{y_2})
\end{align}

We apply lifted case-analysis on each $\pred{S}(\const{N}, \pop{q_j})$. For each $j$, let $\pop{q_{Tj}}$ represent the individuals in $\pop{q_j}$ for which $\pred{S}(\const{N}, \lv{q_j})$ is true and $\pop{q_{Fj}}$ be the other individuals. For each $j$, lifted case-analysis adds two clauses to the theory as follows: 
\begin{align}
\forall \lv{q_j} \in \pop{q_{Tj}}:&~ \pred{S}(\const{N}, \lv{q_j}) \\
\forall \lv{q_j} \in \pop{q_{Fj}}:&~ \neg \pred{S}(\const{N}, \lv{q_j})
\end{align}
We shatter all other singleton clauses based on these newly added singletons. If the singletons are inconsistent, there is no model. Otherwise, let $y_T$ represent $\cup_{j} q_{Tj}$ and $y_F$ represent $\cup_{j} q_{Fj}$. We add the following two singleton clauses to the theory: 
\begin{align}
\forall \lv{y} \in \pop{y_T}:&~ \pred{S}(\const{N}, \lv{y}) \\
\forall \lv{y} \in \pop{y_F}:&~ \neg \pred{S}(\const{N}, \lv{y})
\end{align}
We shatter all clauses having 2S based on these two singletons (not considering the shattering caused by the other singletons) and apply unit propagation. Then the theory will be as follows (the details can be checked in Lemma~\ref{nine-clauses}. Here we only consider the case where $y_T \neq \emptyset$ and $y_F \neq \emptyset$; the case where one of them is empty can be considered similarly as in Lemma~\ref{nine-clauses}):\\

For singletons:
\begin{align}
\forall j: \forall \lv{q_j} \in \pop{q_{Tj}}:&~ \pred{S}(\const{N}, \lv{q_j}) \label{lemma2-r01} \\
\forall j: \forall \lv{q_j} \in \pop{q_{Fj}}:&~ \neg \pred{S}(\const{N}, \lv{q_j}) \label{lemma2-r02}\\
\forall (1, j) \in \psi: \forall \lv{p'_1} \in \pop{p'_1}, \lv{q_j} \in \pop{q_{Tj}}:&~ \pred{S}(\lv{p'_1}, \lv{q_j}) \\
\forall (1, j) \in \psi: \forall \lv{p'_1} \in \pop{p'_1}, \lv{q_j} \in \pop{q_{Fj}}:&~ \pred{S}(\lv{p'_1}, \lv{q_j}) \\
\forall (1, j) \in \overline{\psi}: \forall \lv{p'_1} \in \pop{p'_1}, \lv{q_j} \in \pop{q_{Tj}}:&~ \neg \pred{S}(\lv{p'_1}, \lv{q_j}) \\
\forall (1, j) \in \overline{\psi}: \forall \lv{p'_1} \in \pop{p'_1}, \lv{q_j} \in \pop{q_{Fj}}:&~ \neg \pred{S}(\lv{p'_1}, \lv{q_j}) \\
\forall (i \neq 1, j) \in \psi: \forall \lv{p_i} \in \pop{p_i}, \lv{q_j} \in \pop{q_{Tj}}:&~ \pred{S}(\lv{p_i}, \lv{q_j}) \\
\forall (i \neq 1, j) \in \psi: \forall \lv{p_i} \in \pop{p_i}, \lv{q_j} \in \pop{q_{Fj}}:&~ \pred{S}(\lv{p_i}, \lv{q_j}) \\
\forall (i \neq 1, j) \in \overline{\psi}: \forall \lv{p_i} \in \pop{p_i}, \lv{q_j} \in \pop{q_{Tj}}:&~ \neg \pred{S}(\lv{p_i}, \lv{q_j}) \\
\forall (i \neq 1, j) \in \overline{\psi}: \forall \lv{p_i} \in \pop{p_i}, \lv{q_j} \in \pop{q_{Fj}}:&~ \neg \pred{S}(\lv{p_i}, \lv{q_j}) 
\end{align}

The two singletons on $\pop{y_F}$ and $\pop{y_T}$:
\begin{align}
\forall \lv{y} \in \pop{y_T}:&~ \pred{S}(\const{N}, \lv{y}) \label{lemma2-r03}\\
\forall \lv{y} \in \pop{y_F}:&~ \neg \pred{S}(\const{N}, \lv{y}) \label{lemma2-r04} 
\end{align}

for (1):
\begin{align}
\forall \lv{x} \in \pop{x'}, \lv{y_1}, \lv{y_2} \in \pop{y_T}:&~ \pred{S}(\lv{x}, \lv{y_1}) \vee \pred{S}(\lv{x}, \lv{y_2})\\
\forall \lv{x} \in \pop{x'}, \lv{y_1} \in \pop{y_T}, \lv{y_2} \in \pop{y_F}:&~ \pred{S}(\lv{x}, \lv{y_1}) \vee \pred{S}(\lv{x}, \lv{y_2})\\
\forall \lv{x} \in \pop{x'}, \lv{y_1} \in \pop{y_F}, \lv{y_2} \in \pop{y_T}:&~ \pred{S}(\lv{x}, \lv{y_1}) \vee \pred{S}(\lv{x}, \lv{y_2})\\
\forall \lv{x} \in \pop{x'}, \lv{y_1}, \lv{y_2} \in \pop{y_F}:&~ \pred{S}(\lv{x}, \lv{y_1}) \vee \pred{S}(\lv{x}, \lv{y_2})
\end{align}

for (2): 
\begin{align}
\forall \lv{x} \in \pop{x'}, \lv{y_1}, \lv{y_2} \in \pop{y_T}:&~ \pred{S}(\lv{x}, \lv{y_1}) \vee \neg \pred{S}(\lv{x}, \lv{y_2})\\
\forall \lv{x} \in \pop{x'}, \lv{y_1} \in \pop{y_T}, \lv{y_2} \in  \pop{y_F}:&~ \pred{S}(\lv{x}, \lv{y_1}) \vee \neg \pred{S}(\lv{x}, \lv{y_2})\\
\forall \lv{x} \in \pop{x'}, \lv{y_1} \in \pop{y_F}, \lv{y_2} \in \pop{y_T}:&~ \pred{S}(\lv{x}, \lv{y_1}) \vee \neg \pred{S}(\lv{x}, \lv{y_2})\\
\forall \lv{x} \in \pop{x'}, \lv{y_1} \in \pop{y_F}, \lv{y_2} \in \pop{y_F}:&~ \pred{S}(\lv{x}, \lv{y_1}) \vee \neg \pred{S}(\lv{x}, \lv{y_2})
\end{align}

for (3):
\begin{align}
\forall \lv{x} \in \pop{x'}, \lv{y_1}, \lv{y_2} \in \pop{y_T}:&~ \neg \pred{S}(\lv{x}, \lv{y_1}) \vee \neg \pred{S}(\lv{x}, \lv{y_2})\\
\forall \lv{x} \in \pop{x'}, y_1 \pop{y_T}, \lv{y_2} \in \pop{y_F}:&~ \neg \pred{S}(\lv{x}, \lv{y_1}) \vee \neg \pred{S}(\lv{x}, \lv{y_2})\\
\forall \lv{x} \in \pop{x'}, \lv{y_1} \in \pop{y_F}, \lv{y_2} \in \pop{y_T}:&~ \neg \pred{S}(\lv{x}, \lv{y_1}) \vee \neg \pred{S}(\lv{x}, \lv{y_2})\\
\forall \lv{x} \in \pop{x'}, \lv{y_1} \in \pop{y_F}, \lv{y_2} \in \pop{y_F}:&~ \neg \pred{S}(\lv{x}, \lv{y_1}) \vee \neg \pred{S}(\lv{x}, \lv{y_2})
\end{align}

for (4): 
\begin{align}
\forall \lv{x} \in \pop{x'}, \lv{y} \in \pop{y_F}:&~ \pred{S}(\lv{x}, \lv{y})\\
\forall \lv{x_1}, \lv{x_2} \in \pop{x'}, \lv{y} \in \pop{y_T}:&~ \pred{S}(\lv{x_1}, \lv{y}) \vee \pred{S}(\lv{x_2}, \lv{y})
\end{align}

for (5):
\begin{align}
\forall \lv{x} \in \pop{x'}, \lv{y} \in \pop{y_F}:&~ \neg \pred{S}(\lv{x}, \lv{y})\\
\forall \lv{x} \in \pop{x'}, \lv{y} \in \pop{y_T}:&~ \pred{S}(\lv{x}, \lv{y})
\end{align}

for (6): 
\begin{align}
\forall \lv{x} \in \pop{x'}, \lv{y} \in \pop{y_T}:&~ \neg \pred{S}(\lv{x}, \lv{y})\\
\forall \lv{x_1}, \lv{x_2} \in \pop{x'}, \lv{y} \in \pop{y_F}:&~ \neg \pred{S}(\lv{x_1}, \lv{y}) \vee \neg \pred{S}(\lv{x_2}, \lv{y})
\end{align}

for (7):
\begin{align}
\forall \lv{x} \in \pop{x'}, \lv{y} \in \pop{y_T}:&~ \pred{S}(\lv{x}, \lv{y})\\
\forall \lv{x} \in \pop{x'}, \lv{y} \in \pop{y_F}:&~ \pred{S}(\lv{x}, \lv{y})
\end{align}

for (8): 
\begin{align}
False 
\end{align}

for (9):
\begin{align}
\forall \lv{x} \in \pop{x'}, \lv{y} \in \pop{y_T}:&~ \neg \pred{S}(\lv{x}, \lv{y})\\
\forall \lv{x} \in \pop{x'}, \lv{y} \in \pop{y_F}:&~ \neg \pred{S}(\lv{x}, \lv{y})
\end{align}

Clauses number \ref{lemma2-r01}, \ref{lemma2-r02}, \ref{lemma2-r03} and \ref{lemma2-r04} are disconnected from the rest of the theory and can be reasoned about separately. It is trivial to lift these clauses. Now let's consider the other clauses.

If either one of the clauses \#5, \#7, \#8. or \#9 are in the theory, then unit propagation either gives false or satisfies all clauses. The same is true when both \#4 and \#6 are in the theory. If neither \#4 nor \#6 are in the theory, then we can conjoin the individuals in $\pop{y_T}$ and $\pop{y_F}$ as well as those in $\pop{q_{jT}}$ and $\pop{q_{jF}}$ and write the theory as follows: \\
For singletons:
\begin{align}
\forall (1, j) \in \psi: \forall \lv{p'_1} \in \pop{p'_1}, \lv{q_j} \in \pop{q_{j}}:&~ \pred{S}(\lv{p'_1}, \lv{q_j}) \\
\forall (1, j) \in \overline{\psi}: \forall \lv{p'_1} \in \pop{p'_1}, \lv{q_j} \in \pop{q_{j}}:&~ \neg \pred{S}(\lv{p'_1}, \lv{q_j}) \\
\forall (i \neq 1, j) \in \psi: \forall \lv{p_i} \in \pop{p_i}, \lv{q_j} \in \pop{q_{j}}:&~ \pred{S}(\lv{p_i}, \lv{q_j}) \\
\forall (i \neq 1, j) \in \overline{\psi}: \forall \lv{p_i} \in \pop{p_i}, \lv{q_j} \in \pop{q_{j}}:&~ \neg \pred{S}(\lv{p_i}, \lv{q_j}) 
\end{align}

for (1):
\begin{align}
\forall \lv{x} \in \pop{x'}, \lv{y_1}, \lv{y_2} \in \pop{y}:&~ \pred{S}(\lv{x}, \lv{y_1}) \vee \pred{S}(\lv{x}, \lv{y_2})
\end{align}

for (2): 
\begin{align}
\forall \lv{x} \in \pop{x'}, \lv{y_1}, \lv{y_2} \in \pop{y}:&~ \pred{S}(\lv{x}, \lv{y_1}) \vee \neg \pred{S}(\lv{x}, \lv{y_2})
\end{align}

for (3):
\begin{align}
\forall \lv{x} \in \pop{x'}, \lv{y_1}, \lv{y_2} \in \pop{y}:&~ \neg \pred{S}(\lv{x}, \lv{y_1}) \vee \neg \pred{S}(\lv{x}, \lv{y_2})
\end{align}

This theory is an instance of our initial theory, but with $p_1$ having a smaller domain size. So we can continue this process recursively on the remaining clauses.

Now let's consider the case where \#4 is in the theory but \#6 is not (the case where \#6 is in the theory and \#4 is not is similar). In this case, if \#2 or \#3 are in the theory, then unit propagation either gives false or satisfies all the clauses. If \#4 and \#1 are in the theory, then the theory is as follows: \\

For singletons:
\begin{align}
\forall (1, j) \in \psi: \forall \lv{p'_1} \in \pop{p'_1}, \lv{q_j} \in \pop{q_{Tj}}:&~ \pred{S}(\lv{p'_1}, \lv{q_j}) \\
\forall (1, j) \in \psi: \forall \lv{p'_1} \in \pop{p'_1}, \lv{q_j} \in \pop{q_{Fj}}:&~ \pred{S}(\lv{p'_1}, \lv{q_j}) \\
\forall (1, j) \in \overline{\psi}: \forall \lv{p'_1} \in \pop{p'_1}, \lv{q_j} \in \pop{q_{Tj}}:&~ \neg \pred{S}(\lv{p'_1}, \lv{q_j}) \\
\forall (1, j) \in \overline{\psi}: \forall \lv{p'_1} \in \pop{p'_1}, \lv{q_j} \in \pop{q_{Fj}}:&~ \neg \pred{S}(\lv{p'_1}, \lv{q_j}) \\
\forall (i \neq 1, j) \in \psi: \forall \lv{p_i} \in \pop{p_i}, \lv{q_j} \in \pop{q_{Tj}}:&~ \pred{S}(\lv{p_i}, \lv{q_j}) \\
\forall (i \neq 1, j) \in \psi: \forall \lv{p_i} \in \pop{p_i}, \lv{q_j} \in \pop{q_{Fj}}:&~ \pred{S}(\lv{p_i}, \lv{q_j}) \\
\forall (i \neq 1, j) \in \overline{\psi}: \forall \lv{p_i} \in \pop{p_i}, \lv{q_j} \in \pop{q_{Tj}}:&~ \neg \pred{S}(\lv{p_i}, \lv{q_j}) \\
\forall (i \neq 1, j) \in \overline{\psi}: \forall \lv{p_i} \in \pop{p_i}, \lv{q_j} \in \pop{q_{Fj}}:&~ \neg \pred{S}(\lv{p_i}, \lv{q_j}) \\
\end{align}

for (1):
\begin{align}
\forall \lv{x} \in \pop{x'}, \lv{y_1}, \lv{y_2} \in \pop{y_T}:&~ \pred{S}(\lv{x}, \lv{y_1}) \vee \pred{S}(\lv{x}, \lv{y_2})
\end{align}

for (4): 
\begin{align}
\forall \lv{x} \in \pop{x'}, \lv{y} \in \pop{y_F}:&~ \pred{S}(\lv{x}, \lv{y}) \label{lemma2-r25}\\
\forall \lv{x_1}, \lv{x_2} \in \pop{x'}, \lv{y} \in \pop{y_T}:&~ \pred{S}(\lv{x_1}, \lv{y}) \vee \pred{S}(\lv{x_2}, \lv{y})
\end{align}

Clause number \ref{lemma2-r25} and the singleton clauses having $\lv{q_j} \in \pop{q_{Fj}}$ are disconnected from the rest of the theory and can reasoned about separately. They can be trivially lifted. Once we remove these clauses, the theory will be as follows:

\begin{align}
\forall (1, j) \in \psi: \forall \lv{p'_1} \in \pop{p'_1}, \lv{q_j} \in \pop{q_{Tj}}:&~ \pred{S}(\lv{p'_1}, \lv{q_j}) \\
\forall (1, j) \in \overline{\psi}: \forall \lv{p'_1} \in \pop{p'_1}, \lv{q_j} \in \pop{q_{Tj}}:&~ \neg \pred{S}(\lv{p'_1}, \lv{q_j}) \\
\forall (i \neq 1, j) \in \psi: \forall \lv{p_i} \in \pop{p_i}, \lv{q_j} \in \pop{q_{Tj}}:&~ \pred{S}(\lv{p_i}, \lv{q_j}) \\
\forall (i \neq 1, j) \in \overline{\psi}: \forall \lv{p_i} \in \pop{p_i}, \lv{q_j} \in \pop{q_{Tj}}:&~ \neg \pred{S}(\lv{p_i}, \lv{q_j}) \\
\forall \lv{x} \in \pop{x'}, \lv{y_1}, \lv{y_2} \in \pop{y_T}:&~ \pred{S}(\lv{x}, \lv{y_1}) \vee \pred{S}(\lv{x}, \lv{y_2}) \\
\forall \lv{x_1}, \lv{x_2} \in \pop{x'}, \lv{y} \in \pop{y_T}:&~ \pred{S}(\lv{x_1}, \lv{y}) \vee \pred{S}(\lv{x_2}, \lv{y})
\end{align}
which is an instance of our initial theory, but where $\pop{q_j}$s have smaller domain sizes. So we can continue this process recursively on the remaining clauses.

We showed that in all cases, after domain recursion we will have an instance of our initial theory again, but with smaller domain sizes. By keeping the WFOMC of sub-problems in a cache, the whole process will be domain-liftable: i.e. polynomial in the population sizes.
\end{proof}

\subsection{Proof of Proposition~\ref{subsets-prop}}
\begin{proof}
Let $T \in \FOtwo$ and $T'$ be any of the theories resulting from exhaustively applying rules in \rules\ expect lifted case-analysis on $T$. 
If $T$ initially contains a unary PRV with predicate $\pred{S}$, either it is still unary in $T'$ or lifted decomposition has replaced the LV with a constant. In the first case, we can follow a generic branch of lifted case-analysis on $\pred{S}$, and in the second case, either $T'$ is empty or all binary PRVs in $T$ have become unary in $T'$ due to applying the lifted decomposition and we can follow a generic branch of lifted case-analysis for any of these PRVs. The generic branch in both cases is in $\FOtwo$ and the same procedure can be followed until all theories become empty. If $T$ initially contains only binary PRVs, lifted decomposition applies as the grounding of $T$ is disconnected for each pair of individuals, and after lifted decomposition all PRVs have no LVs. Applying case analysis on all PRVs gives empty theories. Therefore, $T \in RU$. The theory $\forall \lv{x}, \lv{y}, \lv{z} \in \pop{p}: \pred{F}(\lv{x},\lv{y}) \vee \pred{F}(\lv{y},\lv{z}) \vee \pred{F}(\lv{x},\lv{y},\lv{z})$ is an example of a \emph{RU} theory that is not in $\FOtwo$, showing $RU \not\subset \FOtwo$. 
$\FOtwo$ and $RU$ are special cases of $\StwoFOtwo$ and $\StwoRU$ respectively, where $\alpha = \emptyset$, showing $\FOtwo \subset \StwoFOtwo$ and $RU \subset \StwoRU$.
However, Example~\ref{volunteers-jobs} is both in $\StwoFOtwo$ and $\StwoRU$ but is not in $\FOtwo$ and not in \emph{RU}, showing $\StwoFOtwo \not\subset \FOtwo$ and $\StwoRU \not\subset RU$. Since $\FOtwo \subset RU$ and the class of added $\alpha(S)$ clauses are the same, $\StwoFOtwo \subset \StwoRU$.
\end{proof}

\small

\bibliography{MyBib}

\begin{thebibliography}{}

\bibitem[\protect\citeauthoryear{Ahmadi \bgroup \em et al.\egroup
  }{2012}]{ahmadi2012lifted}
Babak Ahmadi, Kristian Kersting, and Sriraam Natarajan.
\newblock Lifted online training of relational models with stochastic gradient
  methods.
\newblock In {\em ECML PKDD}, pages 585--600, 2012.

\bibitem[\protect\citeauthoryear{Ball}{1960}]{birthday-paradox}
W.~W.~Rouse Ball.
\newblock Other questions on probability.
\newblock {\em Mathematical Recreations and Essays}, page~45, 1960.

\bibitem[\protect\citeauthoryear{Beame \bgroup \em et al.\egroup
  }{2015}]{beame2015symmetric}
Paul Beame, Guy {Van den Broeck}, Eric Gribkoff, and Dan Suciu.
\newblock Symmetric weighted first-order model counting.
\newblock In {\em PODS}, pages 313--328, 2015.

\bibitem[\protect\citeauthoryear{Bui \bgroup \em et al.\egroup
  }{2013}]{bui2013automorphism}
Hung~Hai Bui, Tuyen~N Huynh, Artificial~Intelligence Center, and Sebastian
  Riedel.
\newblock Automorphism groups of graphical models and lifted variational
  inference.
\newblock In {\em UAI}, page 132, 2013.

\bibitem[\protect\citeauthoryear{Choi \bgroup \em et al.\egroup
  }{2011}]{Choi:2011}
Jaesik Choi, Rodrigo de~Salvo~Braz, and Hung~H. Bui.
\newblock Efficient methods for lifted inference with aggregate factors.
\newblock In {\em AAAI}, 2011.

\bibitem[\protect\citeauthoryear{Dalvi and Suciu}{2007}]{dalvi2007efficient}
Nilesh Dalvi and Dan Suciu.
\newblock Efficient query evaluation on probabilistic databases.
\newblock {\em The VLDB Journal}, 16(4):523--544, 2007.

\bibitem[\protect\citeauthoryear{{De Raedt} \bgroup \em et al.\egroup
  }{2007}]{DeRaedt:2007}
Luc {De Raedt}, Angelika Kimmig, and Hannu Toivonen.
\newblock {ProbLog}: A probabilistic {P}rolog and its application in link
  discovery.
\newblock In {\em IJCAI}, volume~7, 2007.

\bibitem[\protect\citeauthoryear{{De Raedt} \bgroup \em et al.\egroup
  }{2016}]{StarAI-Book}
Luc {De Raedt}, Kristian Kersting, Sriraam Natarajan, and David Poole.
\newblock Statistical relational artificial intelligence: Logic, probability,
  and computation.
\newblock {\em Synthesis Lectures on Artificial Intelligence and Machine
  Learning}, 10(2):1--189, 2016.

\bibitem[\protect\citeauthoryear{de Salvo~Braz \bgroup \em et al.\egroup
  }{2005}]{De:2005}
Rodrigo de~Salvo~Braz, Eyal Amir, and Dan Roth.
\newblock Lifted first-order probabilistic inference.
\newblock In {\em IJCAI}, pages 1319--1325, 2005.

\bibitem[\protect\citeauthoryear{Getoor and
  Taskar}{2007}]{getoor2007introduction}
Lise Getoor and Ben Taskar.
\newblock {\em Introduction to statistical relational learning}.
\newblock MIT press, 2007.

\bibitem[\protect\citeauthoryear{Gogate and Domingos}{2011}]{PTP}
Vibhav Gogate and Pedro Domingos.
\newblock Probabilistic theorem proving.
\newblock In {\em UAI}, pages 256--265, 2011.

\bibitem[\protect\citeauthoryear{Jaeger}{1997}]{Jaeger:1997}
Manfred Jaeger.
\newblock Relational {B}ayesian networks.
\newblock In {\em UAI}. Morgan Kaufmann Publishers Inc., 1997.

\bibitem[\protect\citeauthoryear{Jernite \bgroup \em et al.\egroup
  }{2015}]{jernite2015fast}
Yacine Jernite, Alexander~M Rush, and David Sontag.
\newblock A fast variational approach for learning {M}arkov random field
  language models.
\newblock In {\em ICML}, 2015.

\bibitem[\protect\citeauthoryear{Jha \bgroup \em et al.\egroup
  }{2010}]{jha2010lifted}
Abhay Jha, Vibhav Gogate, Alexandra Meliou, and Dan Suciu.
\newblock Lifted inference seen from the other side: The tractable features.
\newblock In {\em NIPS}, pages 973--981, 2010.

\bibitem[\protect\citeauthoryear{Kazemi and Poole}{2016a}]{LRC2CPP}
Seyed~Mehran Kazemi and David Poole.
\newblock Knowledge compilation for lifted probabilistic inference: Compiling
  to a low-level language.
\newblock In {\em KR}, 2016.

\bibitem[\protect\citeauthoryear{Kazemi and Poole}{2016b}]{Kazemi:2016}
Seyed~Mehran Kazemi and David Poole.
\newblock Why is compiling lifted inference into a low-level language so
  effective?
\newblock {\em arXiv preprint arXiv:1606.04512}, 2016.

\bibitem[\protect\citeauthoryear{Kersting \bgroup \em et al.\egroup
  }{2009}]{kersting2009counting}
Kristian Kersting, Babak Ahmadi, and Sriraam Natarajan.
\newblock Counting belief propagation.
\newblock In {\em UAI}, pages 277--284, 2009.

\bibitem[\protect\citeauthoryear{Koller and Friedman}{2009}]{Koller:2009}
Daphne Koller and Nir Friedman.
\newblock {\em Probabilistic Graphical Models: Principles and Techniques}.
\newblock MIT Press, Cambridge, MA, 2009.

\bibitem[\protect\citeauthoryear{Kopp \bgroup \em et al.\egroup
  }{2015}]{kopp2015lifted}
Timothy Kopp, Parag Singla, and Henry Kautz.
\newblock Lifted symmetry detection and breaking for {MAP} inference.
\newblock In {\em NIPS}, pages 1315--1323, 2015.

\bibitem[\protect\citeauthoryear{Milch \bgroup \em et al.\egroup
  }{2008}]{Milch:2008}
Brian Milch, Luke~S. Zettlemoyer, Kristian Kersting, Michael Haimes, and
  Leslie~Pack Kaelbling.
\newblock Lifted probabilistic inference with counting formulae.
\newblock In {\em AAAI}, pages 1062--1068, 2008.

\bibitem[\protect\citeauthoryear{Niepert}{2012}]{niepert2012markov}
Mathias Niepert.
\newblock Markov chains on orbits of permutation groups.
\newblock In {\em UAI}, 2012.

\bibitem[\protect\citeauthoryear{Poole \bgroup \em et al.\egroup
  }{2011}]{Poole:2011}
David Poole, Fahiem Bacchus, and Jacek Kisynski.
\newblock Towards completely lifted search-based probabilistic inference.
\newblock {\em arXiv:1107.4035 [cs.AI]}, 2011.

\bibitem[\protect\citeauthoryear{Poole}{2003}]{Poole:2003}
David Poole.
\newblock First-order probabilistic inference.
\newblock In {\em IJCAI}, pages 985--991, 2003.

\bibitem[\protect\citeauthoryear{Richardson and
  Domingos}{2006}]{Richardson:2006aa}
Matthew Richardson and Pedro Domingos.
\newblock {M}arkov logic networks.
\newblock {\em Machine Learning}, 62:107--136, 2006.

\bibitem[\protect\citeauthoryear{Singla and Domingos}{2008}]{singla2008lifted}
Parag Singla and Pedro~M Domingos.
\newblock Lifted first-order belief propagation.
\newblock In {\em AAAI}, volume~8, pages 1094--1099, 2008.

\bibitem[\protect\citeauthoryear{Suciu \bgroup \em et al.\egroup
  }{2011}]{suciu2011probabilistic}
Dan Suciu, Dan Olteanu, Christopher R{\'e}, and Christoph Koch.
\newblock Probabilistic databases.
\newblock {\em Synthesis Lectures on Data Management}, 3(2):1--180, 2011.

\bibitem[\protect\citeauthoryear{Taghipour \bgroup \em et al.\egroup
  }{2013}]{taghipour2013completeness}
Nima Taghipour, Daan Fierens, Guy Van~den Broeck, Jesse Davis, and Hendrik
  Blockeel.
\newblock Completeness results for lifted variable elimination.
\newblock In {\em AISTATS}, pages 572--580, 2013.

\bibitem[\protect\citeauthoryear{{Van den Broeck} \bgroup \em et al.\egroup
  }{2011}]{Van:2011}
Guy {Van den Broeck}, Nima Taghipour, Wannes Meert, Jesse Davis, and Luc {De
  Raedt}.
\newblock Lifted probabilistic inference by first-order knowledge compilation.
\newblock In {\em IJCAI}, pages 2178--2185, 2011.

\bibitem[\protect\citeauthoryear{{Van den Broeck} \bgroup \em et al.\egroup
  }{2012}]{VdBUAI12}
Guy {Van den Broeck}, Arthur Choi, and Adnan Darwiche.
\newblock Lifted relax, compensate and then recover: {F}rom approximate to
  exact lifted probabilistic inference.
\newblock In {\em UAI}, 2012.

\bibitem[\protect\citeauthoryear{{Van den Broeck} \bgroup \em et al.\egroup
  }{2014}]{van2014skolemization}
Guy {Van den Broeck}, Wannes Meert, and Adnan Darwiche.
\newblock Skolemization for weighted first-order model counting.
\newblock In {\em KR}, 2014.

\bibitem[\protect\citeauthoryear{{Van den Broeck}}{2011}]{gvdb2011completeness}
Guy {Van den Broeck}.
\newblock On the completeness of first-order knowledge compilation for lifted
  probabilistic inference.
\newblock In {\em NIPS}, pages 1386--1394, 2011.

\bibitem[\protect\citeauthoryear{{Van Haaren} \bgroup \em et al.\egroup
  }{2015}]{van2015lifted}
Jan {Van Haaren}, Guy {Van den Broeck}, Wannes Meert, and Jesse Davis.
\newblock Lifted generative learning of {M}arkov logic networks.
\newblock {\em Machine Learning}, pages 1--29, 2015.

\bibitem[\protect\citeauthoryear{Venugopal and
  Gogate}{2014a}]{venugopal2014evidence}
Deepak Venugopal and Vibhav Gogate.
\newblock Evidence-based clustering for scalable inference in {M}arkov logic.
\newblock In {\em ECML PKDD}, pages 258--273, 2014.

\bibitem[\protect\citeauthoryear{Venugopal and Gogate}{2014b}]{VenugopalNIPS14}
Deepak Venugopal and Vibhav~G Gogate.
\newblock Scaling-up importance sampling for {M}arkov logic networks.
\newblock In {\em NIPS}, pages 2978--2986, 2014.

\end{thebibliography}
\bibliographystyle{named}

\end{document}